\documentclass[runningheads]{llncs}

\usepackage{eccv}

\usepackage{graphicx}
\usepackage{booktabs}

\usepackage[accsupp]{axessibility}  

\usepackage{multirow}

\begin{document}

\title{THOR: Text to Human-Object Interaction Diffusion via Relation Intervention} 


\author{
\index{Wu, Qianyang}
Qianyang Wu \inst{1}
\and
\index{Shi, Ye}
Ye Shi \inst{1}
\and
\index{Huang, Xiaoshui}
Xiaoshui Huang \inst{2}
\and 
\index{Yu, Jingyi}
Jingyi Yu \inst{1}
\and
\index{Xu, Lan}
Lan Xu \inst{1}
\and
\index{Wang, Jingya}
Jingya Wang \inst{1}
} 
\institute{
ShanghaiTech University  \and
Shanghai AI Laboratory 
}

\maketitle

\begin{abstract}
This paper addresses new methodologies to deal with the challenging task of generating dynamic Human-Object Interactions from textual descriptions (Text2HOI). While most existing works assume interactions with limited body parts or static objects, our task involves addressing the variation in human motion, the diversity of object shapes, and the semantic vagueness of object motion simultaneously. To tackle this, we propose a novel Text-guided Human-Object Interaction diffusion model with Relation Intervention (THOR). THOR is a cohesive diffusion model equipped with a relation intervention mechanism. In each diffusion step, we initiate text-guided human and object motion and then leverage human-object relations to intervene in object motion. This intervention enhances the spatial-temporal relations between humans and objects, with human-centric interaction representation providing additional guidance for synthesizing consistent motion from text. To achieve more reasonable and realistic results, interaction losses is introduced at different levels of motion granularity. Moreover, we construct Text-BEHAVE, a Text2HOI dataset that seamlessly integrates textual descriptions with the currently largest publicly available 3D HOI dataset. Both quantitative and qualitative experiments demonstrate the effectiveness of our proposed model. 

\end{abstract}

\section{Introduction}
\label{sec:intro}
\begin{figure}
    \centering
    \includegraphics[scale=0.48]{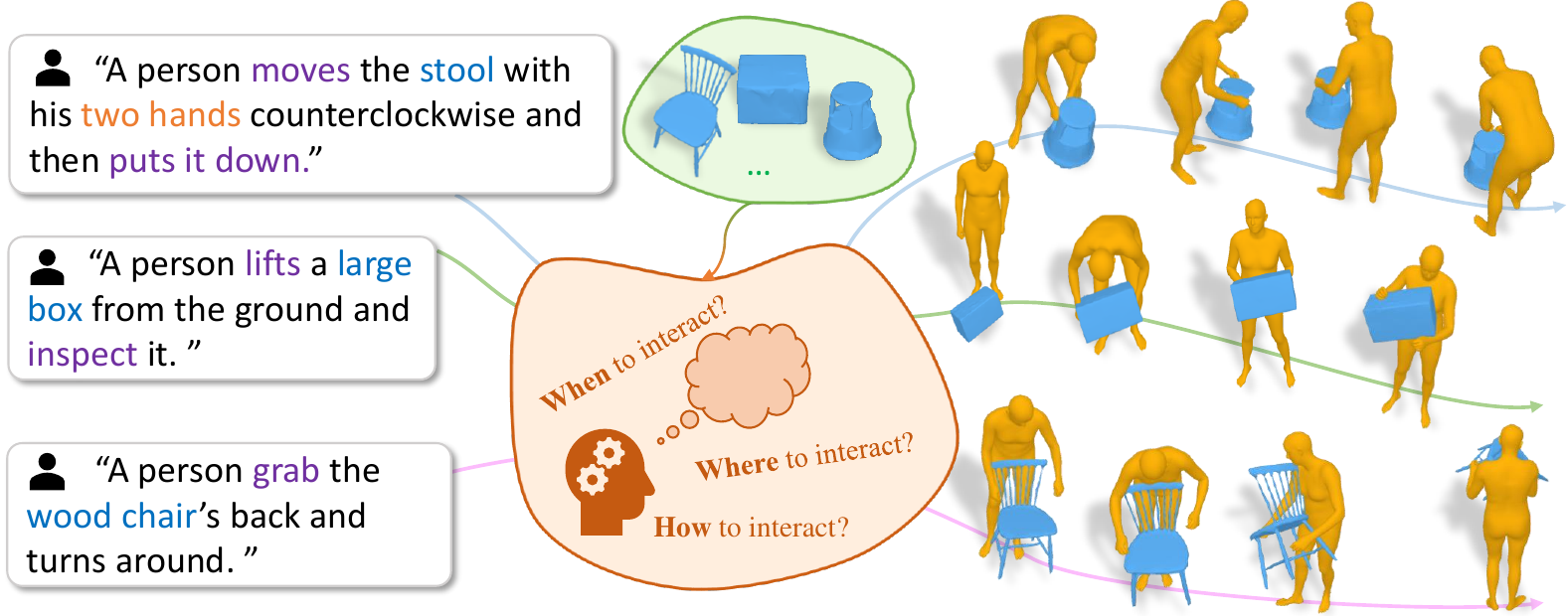}
    \caption{A novel task of generating 3D Human-Object Interaction based on a Text prompt (Text2HOI), reflecting When, Where, and How humans interact with the object.}
    \label{fig:teaser}
\end{figure}

The synthesis of human-object interactions (HOI) is pivotal for various applications, including VR/AR, embodied AI, computer animation, and robotics. Just like humans possess a strong imaginative capability, allowing us to envision interactions with a given object. This imaginative capacity proves beneficial for subsequent planning and execution abilities. Recent advancements in 3D human-object motion capture and generation models have made HOI synthesis increasingly achievable. Given that text serves as a natural and interactive modality for controlling the generation, the imperative need for user-friendly text guided HOI generation method is underscored.

Recently, there have been notable efforts in generating 3D human motions from textual descriptions, showcasing a plausible transition from text to motion \cite{tevet2023MDM, Zhang_2023_remodiffuse, yuan2023physdiff, chen2023mld}. Generating 3D Human-Object Interaction from text (Text2HOI) poses a significant challenge due to the inherent ambiguity in object motion, particularly when considering textual conditions. The ambiguity in object motion introduces complexities, making it challenging to precisely understand and combine the nuanced interactions between humans and objects in three-dimensional space. 

Prior works on generating human-object interactions are either limited with static objects \cite{hassan2021posa, bao2022unidiffuser, pi2023hierarchical, kulkarni2023nifty, zhao2022compositional}, or only generate the human motion of upper body \cite{taheri2021goal, wu2022saga, ghosh2022imos}. 
Generating whole-body interactions with dynamic objects is extremely challenging due to intricate spatial dynamics, the need for accurate trajectory prediction of dynamic objects, and the difficulty in coordinating timing and synchronization.

Taking advantage of the recent advancements in generative models with diffusion \cite{ho2020ddpm, ramesh2022hierarchicaltext, tevet2023MDM}, a straightforward strategy for Text2HOI involves extending existing diffusion models to generate human-object interactions. However, these models encounter challenges in accurately capturing the precise relations between human and object motions. The difficulty arises when explicit contact regions from historical motions \cite{xu2023interdiff} or predefined criteria \cite{li2023OMOMO} are absent, potentially leading to inconsistencies between the object motion and human motion directly generated from text. This discrepancy may stem from the inherent relative structure of joints and their parent joints with fixed bone length, constraining the motion space for humans. Since object dynamics require understanding not only where the interaction occurs but also when and how to interact, textual descriptions fall short in providing such precise contextual information, contributing to the vagueness of the object motion.

To tackle these challenges, we introduce THOR (\textbf{T}ext-conditioned \textbf{H}uman-\textbf{O}bject interaction diffusion with \textbf{R}elation intervention). THOR is a cohesive diffusion model that integrates interaction and intervention mechanisms in a single end-to-end framework.  
As objects play a passive role in interactions, they exhibit a specific spatial distribution determined by their spatial relations with the active human participants. Drawing inspiration from this, we model human-centric relations to refine the object motion.
Specifically, we design the Text2HOI Diffusion with a specially designed Human object relation intervention mechanism that models the human object kinematic relations from the rotation and translation perspectives, respectively. At inference, the intervention still remains in each denoising step. 

To further improve the generation, supervision is introduced on intervened motions, kinematic relations, and geometric distance in a hierarchical manner. To enhance the precise awareness of interactions between human and object, the objective of our model is to denoising not only their motion, but also reasonable kinematic relations between them. While kinematic relations reflect the relative motion pattern between human and object, geometric distance gives point-wise relation for human object interaction. Considering the variation of object shape, it provides fine-grained description of their interactions. These two additional supervision on kinematic relation and geometric distance encourage the generation model to further capture the interaction pattern implicitly.

{\vspace{5pt}
In summary, our contributions are as follows:
\begin{itemize}

\item We propose THOR, a diffusion model specifically tailored for Text2HOI that integrates human-object interactions and intervention mechanisms in a single end-to-end framework. Notably, the intervention mechanism is designed to refine implausible interactions sampled from textual prompts by leveraging the human-object kinematic relations to intervene in the object motion. 

\item We introduce the supervision on human and object kinematic relation and geometric distance to capture multi-level interactions. Through two special objective functions, our model embeds the human object relations into the diffusion process,  facilitating to generate diverse and plausible human object interactions.

\item  We construct a Text2HOI dataset, named Text-BEHAVE that integrates textual descriptions to the currently largest publicly 3D HOI dataset. Both quantitative and qualitative evaluations demonstrate the capability of our model to handle the complexity of this task and produce meaningful and coherent interactions from textual prompts. 

\end{itemize}
}

\begin{figure*}[thp]
  \centering
    \includegraphics[width=\linewidth]{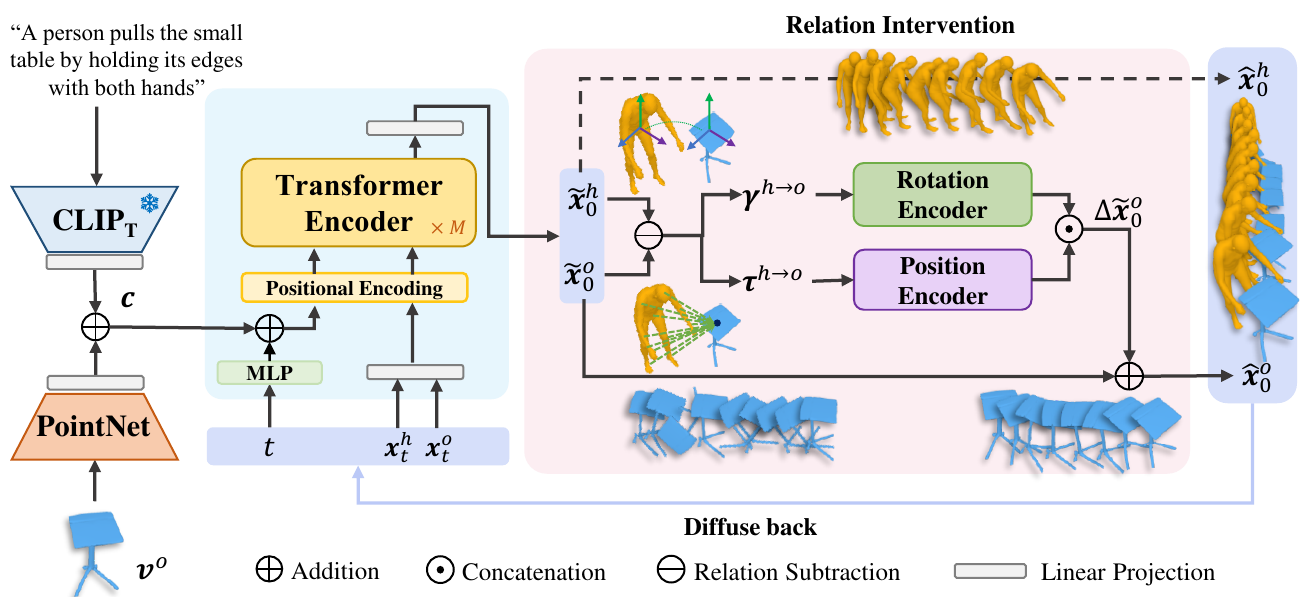}
  \caption{The overview of our model THOR, designed to address a novel task of generating human-object interactions from textual descriptions (Text2HOI). The {\color{black}key innovation} lies in leveraging the human-object spatial relations to further intervene the object motion and then diffuse back to benefit the whole Text2HOI diffusion framework.}
  \label{fig:pipeline}
\end{figure*}
\section{Related Work}

\paragraph{\textbf{Text-to-Human Motion Generation}}

The realm of human motion synthesis has witnessed extensive exploration, with recent endeavors incorporating auxiliary modalities such as audio \cite{li2021choreographer, zhuang2022music2dance, li2022danceformer, alexanderson2023listen} and action \cite{ petrovich2021action, Petrovich_2021_Actor, Xu_2023_actformer} to enhance generation performance and semantic richness. Text-conditioned human motion generation has become a focal point in recent research, with approaches leveraging various techniques including Variational Autoencoders (VAEs) \cite{petrovich2022temos, guo2022T2M, cervantes2022variablelength, athanasiou2022teach}, Vector Quantized VAEs (VQ-VAEs) \cite{lucas2022posegpt, zhong2023attt2m, guo2022tm2t, zhang2023t2mgpt}, and Diffusion models \cite{tevet2023MDM, chen2023mld, Zhang_2023_remodiffuse, yuan2023physdiff, ren2023diffmotion, kim2023flame}. As a powerful pre-trained text-to-image model, CLIP \cite{radford2021CLIP} has been widely adopted for encoding textual descriptions in numerous models \cite{tevet2022motionclip, chen2023mld, Zhang_2023_remodiffuse, Dabral_2023mofusion}. Beyond single-motion generation, efforts have been directed towards long-term human motion synthesis \cite{athanasiou2023sinc, athanasiou2022teach, cao2020longmotionpre, yang2023coherentsamping, gong2023tm2d} and fine-grained motion control on trajectories or joints \cite{shafir2023priorMDM, xie2023omnicontrol, Wang_2023_fgt2m}. Multi-human interactions under text guidance have also been explored, with models like \cite{liang2023intergen} employing cooperative denoising and \cite{Tanaka_2023_RAIG} modeling asymmetric human interactions based on different roles. However, human-only motion usually lack the contextual information as in the real world, this paper we will tackle with the similar text guided generation but involving interactions with objects.

\paragraph{\textbf{Human-Object Interaction}}

In addition to the synthesis of human-only motion, several studies dive into the intricate interactions between humans and objects. Reconstruct human and object interactions from multi-view cameras \cite{bhatnagar2022behave, huang2022intercap, zhang2023neuraldome} and monocular images \cite{zhang2020phosa, xie2022chore, ijcai2023stackflow, xie2023vistracker} has been both investigated. And the generation of human-object interactions has emerged as a prominent focus in recent research. Several approaches leverage human motion priors to generate object positions based on human trajectories and poses \cite{Yi_2022_mover, Ye2022summon, yi2023mime, petrov2023objectpopup, han2023chorus}, providing a human-centric perspective on human-object interactions. In contrast, other works attempts to generate natural human motions in 3D indoor scenes \cite{huang2023scenediffuser, Zhao2023DIMOS} or interactions with seating furniture \cite{chao2021learntosit, zhang2022couch, jiang2023chairs, kulkarni2023nifty, pi2023hierarchical}. These endeavors are further extended by introducing semantic descriptions \cite{hassan2021posa, wang2022humanise, zhao2022compositional, lim2023mammos, xiao2023unihsi}. 

Besides, generating interactions with dynamic objects have been also been investigated, particularly for upper limb interactions with prehensile objects \cite{taheri2021goal, wu2022saga, ghosh2022imos, zhou2022toch, taheri2023grip, razali2023acgmanipulation}, or specific skills involving limited object categories \cite{starke2019nsm, hassan2023physinter, xie2023boxmani, liu2018basketball, merel2020catchcarry, bae2023pmp}. Recently, InterDiff \cite{xu2023interdiff} employs a diffusion model to forecast future interactions from historical observations and OMOMO \cite{li2023OMOMO} builds a two-stage diffusion model to synthesising human motion from moving objects. Also, there are some concurrent works in the mean time. Overall, most existing approaches either focus on static scenes and limited interactions, or lack the text descriptions which is an import interface for generation, leaving a gap in addressing dynamic human-object interactions and the crucial role of textual guidance. Concurrent works for Text2HOI like CG-HOI \cite{diller2023cg} and CHOIS \cite{li2023controllable}, explicitly model human object contact in the diffusion process. HOI-Diff \cite{peng2023hoi} generates HOI in a dual branch diffusion model with affordance guided correction. Without explicit modeling of contact points or affordance, our model interprets interactions based on the kinematic relationship and geometric distance between humans and object surfaces. 

\section{Method}

From a textual description, our goal is to generate plausible human-object interaction. Different from text-to-human motion generation, the object's motion is intricately linked to its shape. Therefore the object shape is also taken as the condition for generation. This problem equals to that, given a text prompt and a 3D object model, generate human-object interactions that satisfy the textual description. This paper combines these two conditional signals and generate human-object interactions in a single diffusion framework. 

We first encode two conditions and generate a primitive result. And this imperfect interaction provides relatively better human motion, which becomes additional guidance to facilitating network to learn object motion.  Sec.~\ref{subsec:framework} will describe the whole diffusion framework that how to encode two conditions and produce primitive interactions. Sec.~\ref{subsec:intervention} explains the details of the novel intervention mechanism and multi-level supervision on interactions, which benefits the whole Text2HOI diffusion framework to refine the primitive generation results by intervening the object motion based on the constructed human-object kinematic relations.

\subsection{Text2HOI Diffusion Framework}
\label{subsec:framework}
\paragraph{\textbf{Motion representation}}
Human and object motion are expected to be in consistent representation. We keep the representation in BEHAVE \cite{bhatnagar2022behave}, where human motion follows the representation in SMPL models \cite{loper2015smpl, MANO:SIGGRAPHASIA:2017, pavlakos2019smplx} and object motion is a series of 6D poses along time dimension. Therefore, human and object are treated as two instances that composed of translations and rotations. In specific, 
human motion and object motion are denoted as 
\begin{equation}
    \boldsymbol x^{h} = [\boldsymbol q, \boldsymbol j], ~~~\boldsymbol x^{o} = [\boldsymbol r, \boldsymbol o],
\end{equation}
where $\boldsymbol q \in \mathbb R^{N\times J\times D_{rot}}$, $\boldsymbol j \in\mathbb R ^{N\times J\times 3}$ refer to joint rotations and joint positions of the human,  $\boldsymbol r \in \mathbb R^{N\times D_{rot}}$ and $\boldsymbol o \in\mathbb R ^{N\times 3}$ are rotations and center translations of object. Here, $N$ is the number of frames, $J$ is the number of human joints, and $D_{rot}$ refer to the dimensions of rotation.  The complete motion is the combination as $\boldsymbol x\triangleq[\boldsymbol x^{h}, \boldsymbol x^{o}]$. Similar to \cite{guo2022T2M, li2023egoego}, human and object motion are canonicalized with the same root orientation and root position in the first frame.   

\paragraph{\textbf{Conditional diffusion}}
Unlike human-only text-to-motion generation, Text2HOI needs to encode both the text prompts and object shape. CLIP \cite{radford2021CLIP} is a powerful pre-trained model that has been verified in many motion generation models \cite{tevet2023MDM, chen2023mld, liang2023intergen}, thus we encode text prompts with the frozen text encoder of CLIP. Similarly, the object shape, noted as $\boldsymbol v_o$ is embedded  by a lightweight shape encoder PointNet \cite{qi2017pointnet}. After respective linear projection, the output text embedding $\boldsymbol c_{text}$ and shape embedding $\boldsymbol c_{shape}$ are added together to obtain the final condition $\boldsymbol c$. 

The complete conditional HOI diffusion framework comprises forward process and reverse process \cite{ho2020ddpm}. The forward process is articulated as $T$ Markov steps, producing noisy interactions $\boldsymbol x_t$ from real interactions $\boldsymbol x_{0}$:
\begin{align}
    q(\boldsymbol x_{1:T}\vert\boldsymbol x_{0}) &= \prod_{t=1}^T q(\boldsymbol x_t \vert \boldsymbol x_{t-1}), \\
    q(\boldsymbol x_t \vert \boldsymbol x_{t-1}) &= \mathcal{N}(\boldsymbol x_t; \sqrt{ \alpha_{t-1}}\boldsymbol x_{t-1}, (1 - \alpha_{t-1})\boldsymbol I),
\end{align}
where $t =1,2,\dots, T$, $\alpha_t \in (0, 1)$ and $\bar \alpha_t = \prod_{i=0}^t\alpha_{i}$.

The reverse process aims to reconstruct realistic interactions $\boldsymbol x_0$ from noisy $\boldsymbol x_T$ and condition $\boldsymbol c$ conposed of text and shape. Following \cite{tevet2023MDM, li2023egoego}, we predict denoised interactions $\boldsymbol x_0$ with network $\mathcal{G}_{\theta}$ in each step and noise it back to $\boldsymbol x_{t-1}$ iteratively following the Markov chain. Denoting $\hat{\boldsymbol x}_0 = \mathcal{G}_{\theta} (\boldsymbol x_t, t, \boldsymbol c)$, the reverse process can be formulated as
\begin{equation}
    p_{\theta}(\boldsymbol x_{t-1} \vert \boldsymbol x_t) = \mathcal{N}(\boldsymbol x_{t-1}; \boldsymbol \mu_{\theta}, \sigma^2\boldsymbol I),
\end{equation}
\begin{equation}
    \boldsymbol{\mu}_{\theta} = \frac{\sqrt{\alpha_t}(1-\bar{\alpha}_{t-1})\boldsymbol x_t + \sqrt{\bar{\alpha}_{t-1}}(1-\alpha_t)\hat{\boldsymbol x}_0}{1-\bar{\alpha}_t},
\end{equation}
where $\sigma^2$ is a fixed variance.

\paragraph{\textbf{Classifier-free guidance}}
Previous diffusion models \cite{ramesh2022hierarchicaltext, tevet2023MDM, chen2023mld, liang2023intergen} achieve classifier-free guidance by the linear combination of unconditional generation and conditional generation. However, in the context of Text2HOI generation, object shape has to serve as an essential condition modality. This is mainly because, even within the same class, objects with different shapes possess distinguishable motion spaces. This distinction may arise from variations in size or geometric structure, impacting how objects move or interact. Consequently, the classifier-free guidance is reformulated as:
\begin{equation}
    \mathcal{G}_{\theta}(\boldsymbol x_t, \boldsymbol c, t) = (1-s)\mathcal{G}_{\theta}(\boldsymbol x_t, \boldsymbol c_{shape}, t) + s\mathcal{G}_{\theta}(\boldsymbol x_t, \boldsymbol c, t),
\end{equation}
where $s$ is the guidance scale. With this modification, THOR maintains a consistent awareness of the object's shape, striking a balance between fidelity and diversity guided by the provided text prompt.

In each diffusion step $t$, the noisy interactions $\boldsymbol x_t$ is linearly projected, and concatenated with the condition $\boldsymbol c$. We use a transformer encoder with $M$ layers to predict the denoised $\Tilde{\boldsymbol x}_0$ with positional encoding. This builds a general Text2HOI generation model if directly go to next step. However, such a simple diffusion model tends to learn the entire distribution of input data, while lacking the nuanced awareness regarding interactions. Our model, THOR, addresses this limitation by introducing human object relation intervention as in the next subsection.

\subsection{Diffusion with Human-Object Relation Intervention }
\label{subsec:intervention}

\begin{figure*}[t]
  \centering
    \includegraphics[width=\linewidth]{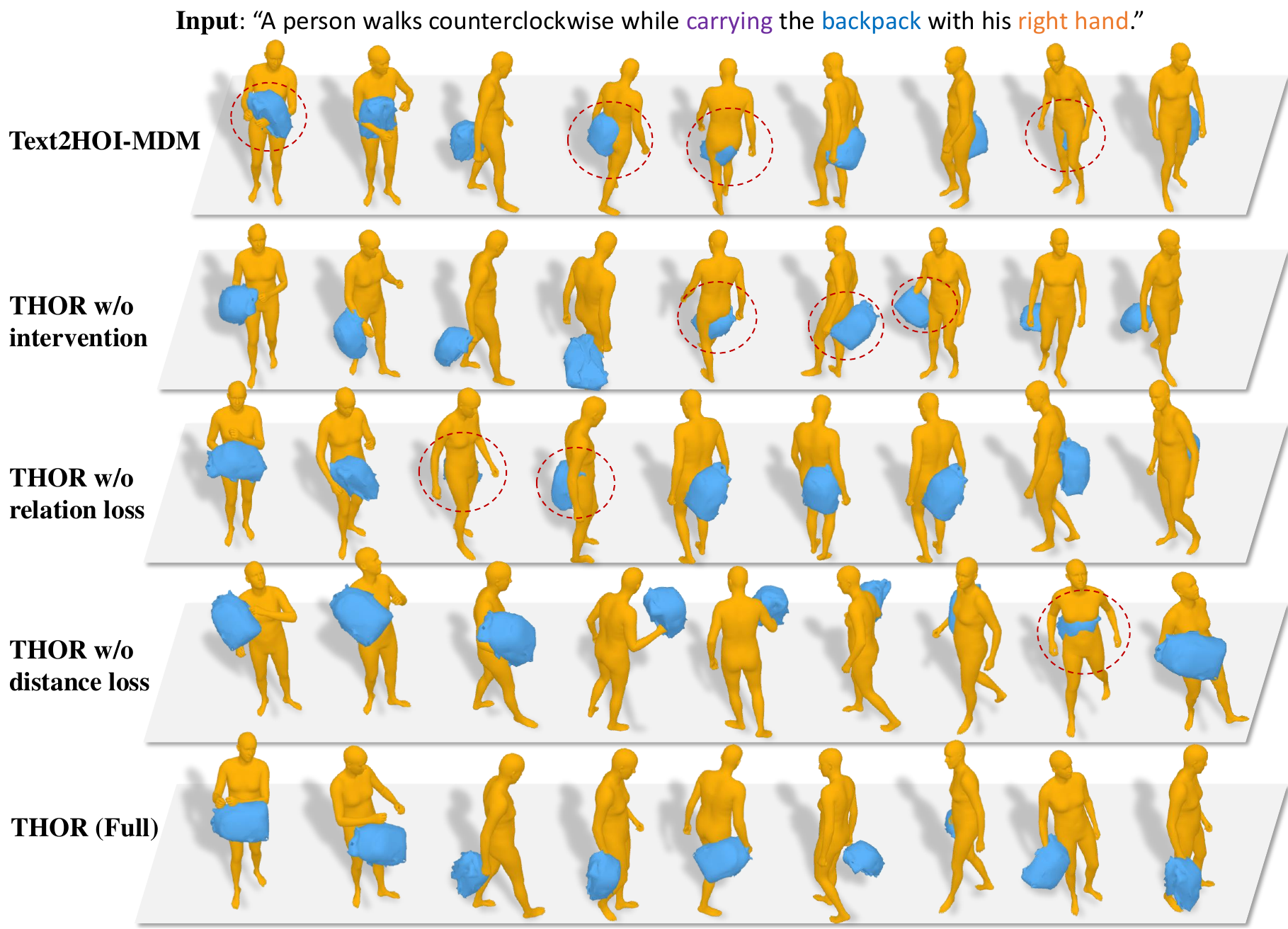}
  \caption{Qualitative comparison for Text2HOI generation. Artifacts are highlighted in red circle. Our model, THOR, can generate realistic and plausible human-object interaction in response to the textual guidance. Through the intervention mechanism and two interaction losses, it corrects the drifting object motion and alleviate the implausible human-object spatial relations.}
  \label{fig:comparison}
\end{figure*}
As highlighted in Sec.~\ref{sec:intro}, text is a weak condition on object motion, which can lead to deficiencies in the overall interaction generation if object motion is directly sampled from it. Given object shape and coarse human motion, human ourselves can infer and envision plausible object motion in response. Acknowledging this, the human motion can serve as auxiliary guidance to enhance semantic meanings of interactions learned from text, providing a more intuitive and contextually rich perspective in generation.

\paragraph{\textbf{Modeling Human-object Kinematic Relations}}
To bridge the gap between human motion and object motion in interaction, we introduce a novel intervention network designed to rectify implausible object motion. Rather than predicting object motion directly from human motion, which might neglect the initial learned interaction, our approach leverages their spatial relations to predict residual correction terms on object motion. These relations encapsulates the object's rotations and translations relative to each human joint from a human-centric perspective. 

\begin{equation}
\setlength\abovedisplayskip{5pt}\vspace{5pt}
\setlength\belowdisplayskip{5pt}
        \boldsymbol \gamma^{h\rightarrow o} = \boldsymbol q \ominus \boldsymbol r,~~~~
        \boldsymbol \tau^{h\rightarrow o} = \boldsymbol j \ominus \boldsymbol o.
\end{equation}
Here, $\boldsymbol{\gamma}^{h\rightarrow o} \in \mathbb{R}^{N\times J\times 3}$ represents the rotation relations, $\boldsymbol{\tau}^{h\rightarrow o} \in \mathbb{R}^{N\times J\times 3}$ represents translation relations, and $\ominus$ denotes human joint-wise subtraction. At the $t$-th step of diffusion model, the primitive motion $\Tilde{\boldsymbol x}^h_0$ and $\Tilde{\boldsymbol x}^o_0$ would be reformulated as $\Tilde{\boldsymbol \gamma}^{h\rightarrow o}$ and $\Tilde{\boldsymbol \tau}^{h\rightarrow o}$ similarly.

This unique human-centric relation representation enhances the model's understanding of the relationships between the human body and objects. By concentrating on object motion directly associated with human body poses, our approach not only improves the model's contextual awareness but also lays the groundwork for generating realistic, consistent, and semantically meaningful human-object interactions. 

\paragraph{\textbf{Relation Intervention Network}}
Recognizing that rotations and translations are distinct transformations, we intervene separately to guarantee that the output residual intervention preserve a scale aligned with the input relation. After jointly modeling of human and object motion with two transformation representation, separating them can enhance the awareness of distinction between these two motion components. The intervention network is composed of two lightweight transformer encoders, named \textbf{{Rotation encoder}} and \textbf{{Position encoder} }respectively. Rotation relations and translation relations are individually input to these encoders and then we predict two residual terms $\Delta \Tilde{\boldsymbol r}$ and $ \Delta \Tilde{\boldsymbol o}$. Consequently, we obtain a perturbation term on object motion, $\Delta \Tilde{\boldsymbol x}^o_0 = [\Delta \Tilde{\boldsymbol r}, \Delta \Tilde{\boldsymbol o}]$. Through the attention mechanism in transformer, it aggregates intervention from each human joint to improve the object motion. Finally, this $\Delta \Tilde{\boldsymbol x}^o_0$ is added back to the primitive object motion $\Tilde{\boldsymbol x}^o_0$ to obtain the final estimated $\hat {\boldsymbol x}_0^o$.  

Benefited from the cyclical nature of the diffusion model, early rectification of object motion not only enhances the object motion but also contributes to the refinement of human motion throughout the diffusion process. This seamless integration eliminates the need for additional fusion operations, emphasizing the efficiency and coherence of our approach.We take both human motion and object motion as input to model their joint distribution. The simple objective \cite{ho2020ddpm, tevet2023MDM} for our diffusion framework is
\begin{equation}
        \mathcal{L}_{simple} =  \mathbb{E}_{t\sim[1, T]} \Vert \hat{\boldsymbol{x}}_{0}-{\boldsymbol{x}}_{0}\Vert.
\end{equation}

\paragraph{\textbf{Interaction Losses}}
We present two interaction losses to supervise the kinematic relation and geometric distance between human and object to further enhance the generation ability. The first relation loss aims to supervise the kinematic relations to encourage interaction network to generate reasonable relations. Then the model is guided to generate interactions that not only adhere to textual descriptions but also align with reliable spatial configurations of human and object. 

This enhances the overall realism of the generated human-object interactions. The relation loss is formulated as:
\begin{equation}
    \mathcal{L}_{rel} = \mathbb{E}_{t\sim[1, T]} (\Vert \hat{\boldsymbol{\gamma}}^{h\rightarrow o}_{0} - \boldsymbol{\gamma}^{h\rightarrow o}_{0} \Vert + \Vert \hat{\boldsymbol{\tau}}^{h\rightarrow o}_{0} - {\boldsymbol{\tau}}^{h\rightarrow o}_{0} \Vert ).
    \vspace{6pt}
\end{equation}
However, relation loss do not explicitly take the object shape into consideration, which do not modeling the geometric transformation relationship between object pose and points on its surface, and this may cause severe penetrations. Instead, the distance field between human joints and object surface can provide finer-grained geometry cues for HOI. The explicit distance restricts the object spatial distribution around human body by considering its shape. Therefore, we reconstruct their surface and calculate the signed distance between human joints and object points. Specifically, the mutual signed distance is represented as $\hat{\boldsymbol{d}}^{h\rightarrow o}_{0} \in \mathbb{R}^{N\times M\times 3}$ and $\hat{\boldsymbol{d}}^{o\rightarrow h}_{0} \in \mathbb{R}^{N\times M\times 3}$, And the distance loss can be written as: 
\begin{equation}
\mathcal{L}_{dist} \!= \mathbb{E}_{t\sim[1, t']} ( \Vert \hat{\boldsymbol{d}}^{h\rightarrow o}_{0}\!-\boldsymbol{d}^{h\rightarrow o}_{0} \Vert+ \Vert \hat{\boldsymbol{d}}^{o\rightarrow h}_{0}\!-\boldsymbol{d}^{o\rightarrow h}_{0} \Vert ).
\vspace{10pt}
\end{equation}
Similar to \cite{yuan2023physdiff,liang2023intergen, xu2023interdiff}, this is applied in only few diffusion steps under a threshold $t'$.

With the simple objective mentioned before, our approach supervise the interactions at multiple levels, which enhances the model's capacity to discern and understand interactions in a progressive way, providing a more comprehensive supervision of the underlying dynamics. The total training loss of our diffusion is summarized as:
\begin{equation}
    \mathcal{L} = \mathcal{L}_{simple} + \lambda_{rel}\mathcal{L}_{rel} + \lambda_{dist}\mathcal{L}_{dist} + \lambda_{vel}\mathcal{L}_{vel},
\end{equation}
where $\mathcal{L}_{vel} =\mathbb{E}_{t\sim[1, T]}\Vert \hat{\boldsymbol x}_{0, 1:N} - \hat{\boldsymbol x}_{0, 0:N-1} \Vert$ represents velocity regularization.

\section{Our Text-BEHAVE Dataset }
We have enriched the BEHAVE dataset, currently the largest publicly accessible dataset for 3D Human-Object Interaction \cite{bhatnagar2022behave}. This enrichment involves the manual annotation of textual descriptions for each sequence.
Specifically, each sequence is partitioned into multiple clips based on the annotations, ensuring semantic consistency throughout every clip. 

Besides, we discard clips that are meaningless or too short and randomly crop the clips with duration longer than 10s. This process results in a text annotated human-object interaction dataset, including total of 2377 interaction clips ranging from 2 to 10s. The total interaction length amounts 440, 840 frames at 30 fps. It contains 18 objects belonging to 12 categories (e.g. box, table, chair, backpack), and various common interactions (e.g. `lifting', `sitting', `dragging') with these objects. The average length of textual descriptions is 19.7, which includes consecutive interactions with conjunction. As for interaction representation, human motion follows the format in SMPL-H \cite{MANO:SIGGRAPHASIA:2017} and object is represented in 6D pose including rotation and translation. In total, there are 2144 clips for training and 233 clips for testing, respectively. For brevity, we refer to this augmented dataset as \textbf{Text-BEHAVE}. 

\section{Experiments}

\paragraph{\textbf{Datasets and Metrics}}
Our model are trained and tested on the Text-BEHAVE. As in \cite{guo2022T2M}, we evaluate the generation ability of our model with following metrics. \emph{FID} evaluates the difference of distribution between the generated motion and real motion. \emph{Diversity} evaluates the dissimilarity between generated motions, while \emph{MModality} measures it within the same condition. \emph{R precision} assesses the consistency between text and the generated motion in a retrieval way, \emph{MM Dist} measures the feature distance between conditions and generated motions. We use contrastive loss to train a text encoder and motion encoder on the extended Text-BEHAVE dataset.  Additionally, we also adopt the motion reconstruction metrics from \cite{bhatnagar2022behave, zhang2023neuraldome, li2023OMOMO}. We sample 20 times from input text and the best results are selected, as in \cite{li2023OMOMO, xu2023interdiff}. Furthermore, user study is introduced to evaluate the visual perceptual performance.

\paragraph{\textbf{Baselines}}
We compare our model with a VAE model whose encoder and decoder are from \cite{petrovich2022temos}, and a motion diffusion model from \cite{tevet2023MDM}. They are modified to suit Text2HOI task, noted as `Text2HOI-VAE' and `Text2HOI-MDM' respectively.

\paragraph{\textbf{Implementation details}}
 We use Adam optimizer and learning rate $10^{-4}$. The batch size is set to 64 and the number of diffusion steps $T$ is $1000$. For the text encoder, we choose the CLIP pre-trained text encoder in the version of `ViT-B/32'. The text embedding is randomly masked by a ratio 0.1 for classifier-free guidance. The primitive generation uses a transformer encoder of 8 layers and intervention network comprises of 2 transformer layers. All the experiments were implemented on a single NVIDIA GeForce RTX 4090 with 24 GB of memory, and our model takes about 20 hours to converge.
\subsection{Quantitative Results}

As shown in Table \ref{tab:generation}, our approach outperforms the baselines across Text2HOI generation metrics. Through the intervention mechanism, our model can enhance the text guidance by the preliminary generated human motion, which further improve the performance on the retrieval-based metrics and multi-modality metrics. Additionally, THOR exhibits a significantly lower \emph{FID}  than Text2HOI-MDM, indicating a notable improvement in generating high-quality motions comparable to real ones. Our method also excels in terms of \emph{diversity}, showcasing a more varied range of generated motions.This is crucial as it suggests that our approach can provide a broader range of human motions, contributing to more varied and realistic results. At last, the \emph{MModality} score highlights the richness of multi-modality in our generated motions.

\begin{table}[h]
	\begin{center}
		\centering
		\caption{\textbf{Quantitative comparisons on the Text-BEHAVE test set for Text2HOI generation.} All the evaluations run 20 times. $\pm$ indicates the 95\% confidence interval. \textbf{Bold} indicates best result.} \label{tab:generation}
\resizebox{\textwidth}{!}{		
		\begin{tabular}{lccccccc}
			\toprule
			{~~~~~Methods}  & \multicolumn{3}{c}{R Precision$\uparrow$} & {FID $\downarrow$} & {MM Dist$\downarrow$}  & {Diversity$\rightarrow $} & {MModality $\uparrow$} \\
			\cmidrule(lr){2-4}
			& Top 1 & Top 2  & Top 3 \\
			\midrule
			\textbf{Real motions}    &  $0.994^{\pm .003}$ &  $0.998^{\pm .001}$  & $1.000^{\pm .000}$   &  $0.031^{\pm .002}$ &  $3.377^{\pm .004}$  & $22.832^{\pm .074}$   & $-$  \\
			\midrule
			Text2HOI-VAE  &  $0.095^{\pm .008}$ &  $0.142^{\pm .011}$  & $0.178^{\pm .016}$   &  $4.274^{\pm .032}$ &  $8.074^{\pm .018}$  & $17.121^{\pm .134}$   & $1.612^{\pm .123}$  \\
			Text2HOI-MDM   &  $0.186^{\pm .012}$ &  $0.286^{\pm .010}$  & $0.348^{\pm .007}$   &   $2.680^{\pm .028}$&  $7.065^{\pm .014}$  & ${22.376}^{\pm .088}$   & $2.261^{\pm .079}$ \\
			\midrule
   			THOR w/o intervention   &  $0.194^{\pm .007}$ &  $0.306^{\pm .011}$  & $0.387^{\pm .015}$   &  $1.909^{\pm .015}$ &  $7.008^{\pm .018}$  & $23.015^{\pm .141}$ & $2.247^{\pm .109}$ \\
         	THOR w/o dist. loss   &  $0.198^{\pm .007}$ &  $0.314^{\pm .008}$  & $0.390^{\pm .015}$   &  $1.936^{\pm .018}$ &  $6.933^{\pm .010}$  & $23.025^{\pm .086}$ & $2.393^{\pm .053}$ \\
         	THOR w/o rel. loss   &  $0.218^{\pm .005}$ &  $0.338^{\pm .008}$  & $\textbf{0.420}^{\pm .008}$  &  $\textbf{1.874}^{\pm .022}$ &  $6.887^{\pm .011}$  & $22.475^{\pm .077}$ & $2.093^{\pm .070}$ \\
			THOR (Full)    &  $\textbf{0.250}^{\pm .009}$ &  $\textbf{0.362}^{\pm .008}$  & $0.411^{\pm .001}$   &  $1.983^{\pm .026}$ &  $\textbf{6.874}^{\pm .014}$  & $\textbf{22.701}^{\pm .081}$ & $\textbf{2.575}^{\pm .071}$ \\
			\bottomrule
		\end{tabular}
  }
	\end{center}
\end{table}

\subsection{Qualitative Results}

To demonstrate the visual performance of THOR, we provide qualitative results for evaluation. Text2HOI-VAE is excluded since it generate implausible results. As shown in Figure \ref{fig:comparison}, we give an example text prompt and the generated interactions. To demonstrate the effectiveness of our model, we highlight the artifacts in other alternatives. In the absence of intervention, models may produce the drifting object or interactions that is misaligned with textual descriptions. And without relation loss and distance loss, generated interactions may encounter penetration or fail to establish contact with the human body. However, our model demonstrates the capability to generate consistent and realistic human-object interactions, ensuring the object motion aligns with the human motion through the intervention mechanism and interaction losses.

\begin{figure}[t]
  \centering
   \includegraphics[width=0.85\linewidth]{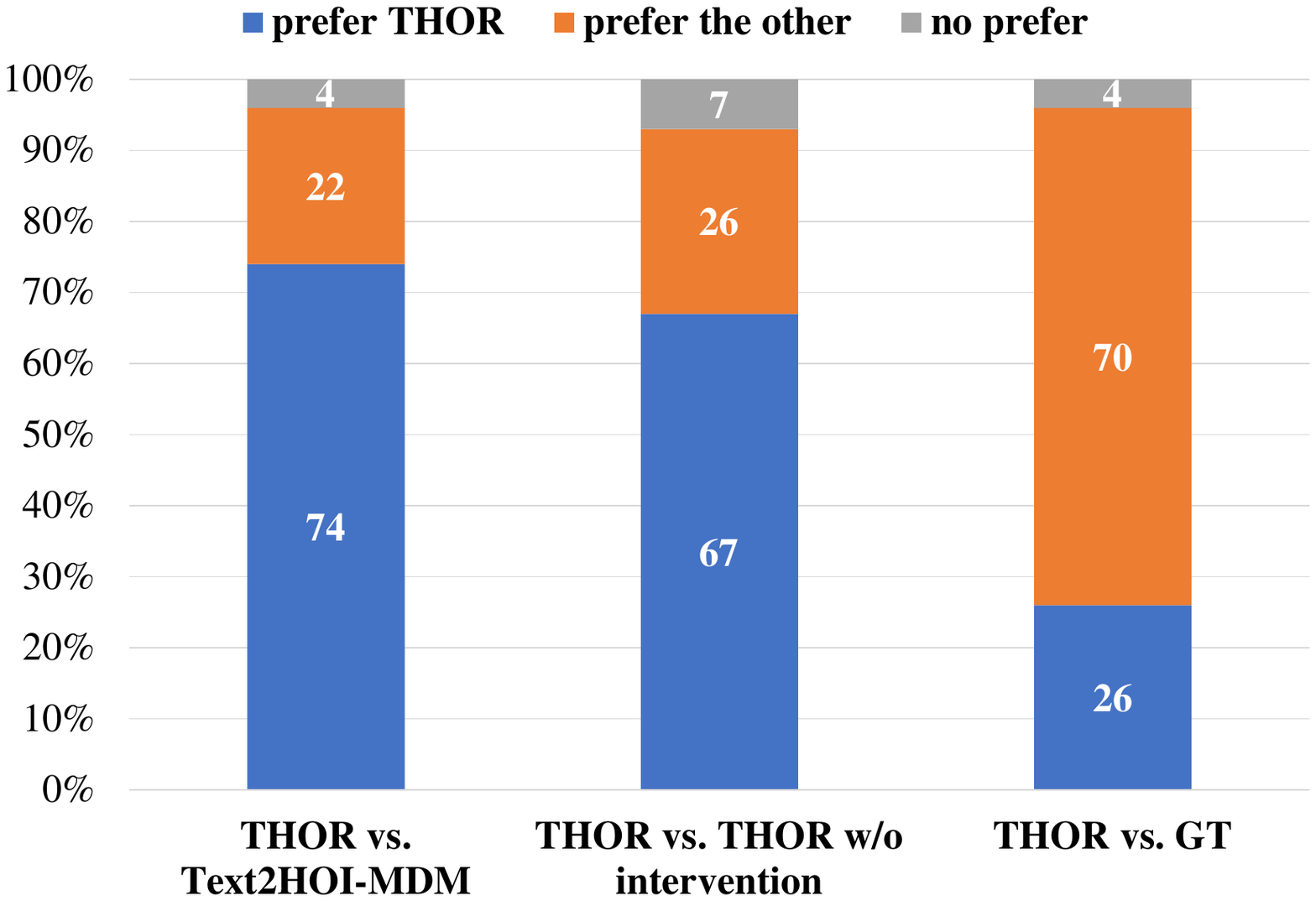}

   \caption{User study on Text-BEHAVE test set, where `w/o' indicates `without' and `GT' indicates ground truth.}
   \label{fig:userstudy}
\end{figure}

The user study is also conducted, serving as a complementary evaluation. We carefully selected 30 samples, encompassing 10 object classes, each generated from 3 text prompts for every object. Subsequently, we invited 27 users to assess the generation quality and express their preferences between two choices. The results, depicted in Figure \ref{fig:userstudy}, demonstrate that our model outperforms others and is comparable to the ground truth in certain samples.
\begin{figure*}[htbp]
  \centering
   \includegraphics[width=\linewidth]{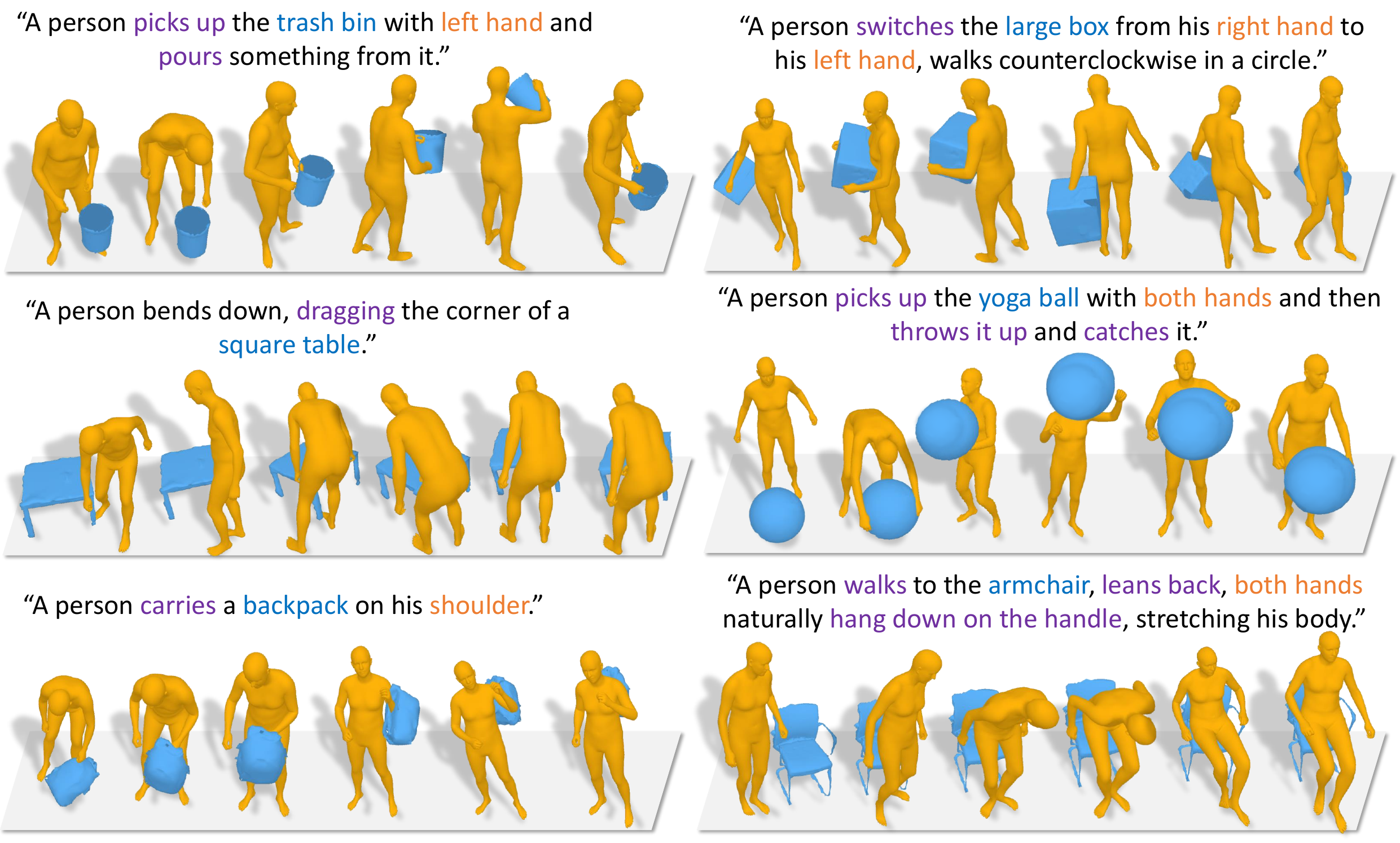}
   \caption{\textbf{Qualitative results of our model for Text2HOI generation.} Our model can generate human-object interactions aligned with the text description involving static and dynamic objects with diverse categories and shapes.}
   \label{fig:onecol}
\end{figure*}

\begin{figure}[htbp]
    \centering
    \begin{subfigure}{0.49\textwidth}
        \includegraphics[width=\linewidth]{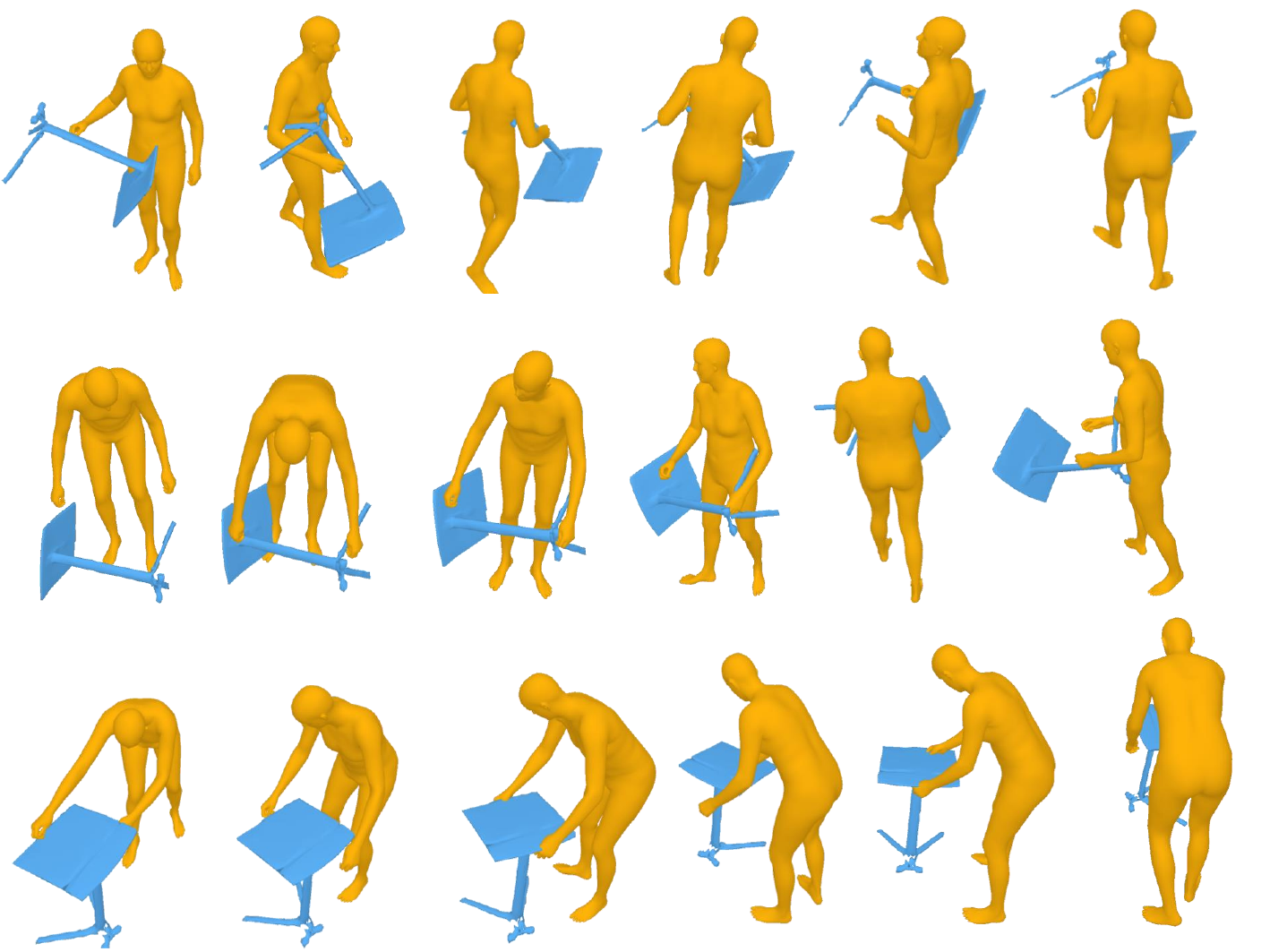}

        \caption{Generate multiple interactions with the same table.}
        \label{fig:gs0_table}
    \end{subfigure}%
    \vspace{10pt}
    \centering
    \begin{subfigure}{0.49\textwidth}  
        \includegraphics[width=\linewidth]{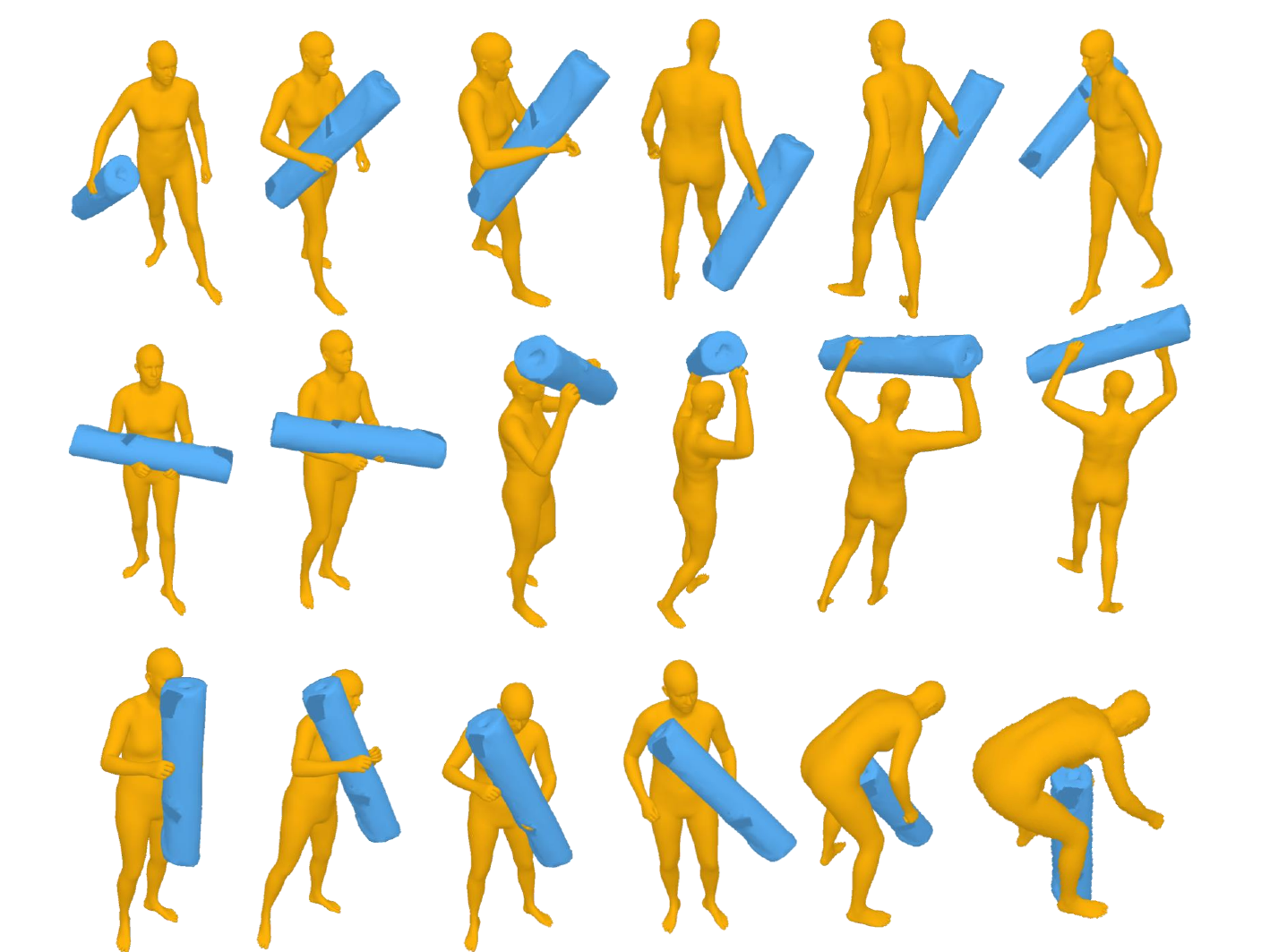}

        \caption{Generate multiple interactions with the same yoga mat.}
        \label{fig:gs0_yogamat}
    \end{subfigure}
    \caption{Sampling interactions from only object shape. It shows that our model can generate plausible interactions in response to a given object shape.}
    \label{fig:gs0}
\end{figure}

If guidance scale $s=0$ in sampling,  the model turns to generate interactions solely conditioned on a object model. As shown in Figure~\ref{fig:gs0}, without textual guidance, our model can generate human-object interactions satisfying different object shape. This also verifies the effectiveness of the modified HOI classifier-free guidance, which strikes a balance for the fidelity matching the textual description and diversity with respect to the object geometry.

\subsection{Ablation study}

To better understand the contribution of each component in our proposed method, THOR, we conducted ablation studies by removing specific elements from the model architecture. The quantitative results are summarized in Table \ref{tab:generation}, and we discuss them as below:
Our THOR model achieves the best overall performance. 
The absence of intervention mechanism causes a distinct decrease in \emph{R Precision}, \emph{MM Dist}, and \emph{MModality}, confirming that the intervention from human motion guidance ensures a more precise alignment with textual guidance. The comparable \emph{FID} and \emph{Diversity} scores indicate that the model retains its ability to generate diverse and plausible motions. The distance loss results in a minor decrease in \emph{R Precision} and \emph{Diversity}, indicating that this loss component contributes to the accuracy and variety of generated motions. Surprisingly, the \emph{FID} score improves, suggesting that without distance loss, the model generates motions that are closer to real motions but potentially sacrifices diversity. Excluding the relation loss slightly reduces accuracy but leads to improved visual fidelity, suggesting that relation loss may introduce certain variations in generated motions. Applying these two interaction losses provides additional guidance to the model. By supervising the relations in the diffusion steps, the model is encouraged to generate reasonable spatial relations from the outset, contributing to the overall coherence of the generated interactions.  This confirms that the combination of all proposed components leads to a balanced generation of fidelity, diversity guided by textual input. 

Qualitative ablation studies are illustrated in Figure \ref{fig:comparison}, highlighting the proficiency of our complete model in generating interactions with superior visual performance. The intervention mechanism and the relation loss addresses issues such as drifting object motion and meaningless trajectories, while the distance loss effectively prevents penetration.  The ablation studies reveal that each component in THOR contributes to specific aspects of motion generation. The intervention mechanism and relation loss ensures precise and diverse motion synthesis, while distance loss improves the visual fidelity. THOR achieves a synergistic effect, striking a balance between accuracy and diversity in text-guided interaction generation.

\section{Conclusion}

In summary, we proposed a novel diffusion model to deal with Text2HOI, addressing the challenge of generating full-body interactions with dynamic objects guided by textual prompts. This task involves the simultaneous generation of consistent human and object motion, posing a significant difficulty. To address this, we present THOR, a diffusion framework seamlessly integrating human-object interaction and intervention mechanisms within a unified end-to-end framework. We interprets interactions based on the kinematic relation and geometric distance between humans and object
surfaces, introducing two interaction losses to generate plausible and realistic results.  
Additionally, we contribute the Text-BEHAVE dataset, integrating textual descriptions into the largest publicly available 3D HOI dataset. Both quantitative and qualitative experiments are conducted to validate the effectiveness of THOR in generating coherent and realistic human-object interactions. 

\paragraph{\textbf{Limitations and future works}} Though our model generates realistic results, issues like penetration and floating still exists in some generated results, especially for objects with complex shape and unseen objects. The existing HOI datasets fall short in both scale and quality compared to human motion datasets, highlighting the clear need for more extensive and high-quality HOI data.  The expansion of these datasets would be highly valuable to advance research in human-object interactions. Additionally, future works could benefit from integrating dexterous hand motion to generate more comprehensive human-object interactions. The exploration of long-term generation and fine-granularity control remains an area for further investigation.

\newpage

\bibliographystyle{splncs04}

\bibliography{main}

\begin{thebibliography}{10}
\providecommand{\url}[1]{\texttt{#1}}
\providecommand{\urlprefix}{URL }
\providecommand{\doi}[1]{https://doi.org/#1}

\bibitem{alexanderson2023listen}
Alexanderson, S., Nagy, R., Beskow, J., Henter, G.E.: Listen, denoise, action!
  audio-driven motion synthesis with diffusion models. ACM Transactions on
  Graphics (TOG)  \textbf{42}(4),  1--20 (2023)

\bibitem{athanasiou2022teach}
Athanasiou, N., Petrovich, M., Black, M.J., Varol, G.: Teach: Temporal action
  composition for 3{D} humans. In: 2022 International Conference on 3D Vision
  (3DV). pp. 414--423. IEEE (2022)

\bibitem{athanasiou2023sinc}
Athanasiou, N., Petrovich, M., Black, M.J., Varol, G.: Sinc: Spatial
  composition of 3{D} human motions for simultaneous action generation. arXiv
  preprint arXiv:2304.10417  (2023)

\bibitem{bae2023pmp}
Bae, J., Won, J., Lim, D., Min, C.H., Kim, Y.M.: Pmp: Learning to physically
  interact with environments using part-wise motion priors. arXiv preprint
  arXiv:2305.03249  (2023)

\bibitem{bao2022unidiffuser}
Bao, F., Nie, S., Xue, K., Li, C., Pu, S., Wang, Y., Yue, G., Cao, Y., Su, H.,
  Zhu, J.: One transformer fits all distributions in multi-modal diffusion at
  scale. In: International Conference on Machine Learning. pp. 1692--1717. PMLR
  (2023)

\bibitem{bhatnagar2022behave}
Bhatnagar, B.L., Xie, X., Petrov, I.A., Sminchisescu, C., Theobalt, C.,
  Pons-Moll, G.: Behave: Dataset and method for tracking human object
  interactions. In: Proceedings of the IEEE/CVF Conference on Computer Vision
  and Pattern Recognition. pp. 15935--15946 (2022)

\bibitem{bhattacharyya2018bestofmany}
Bhattacharyya, A., Schiele, B., Fritz, M.: Accurate and diverse sampling of
  sequences based on a “best of many” sample objective. In: Proceedings of
  the IEEE Conference on Computer Vision and Pattern Recognition. pp.
  8485--8493 (2018)

\bibitem{cao2020longmotionpre}
Cao, Z., Gao, H., Mangalam, K., Cai, Q.Z., Vo, M., Malik, J.: Long-term human
  motion prediction with scene context. In: Computer Vision--ECCV 2020: 16th
  European Conference, Glasgow, UK, August 23--28, 2020, Proceedings, Part I
  16. pp. 387--404. Springer (2020)

\bibitem{cervantes2022variablelength}
Cervantes, P., Sekikawa, Y., Sato, I., Shinoda, K.: Implicit neural
  representations for variable length human motion generation. In: European
  Conference on Computer Vision. pp. 356--372. Springer (2022)

\bibitem{chao2021learntosit}
Chao, Y.W., Yang, J., Chen, W., Deng, J.: Learning to sit: Synthesizing
  human-chair interactions via hierarchical control. In: Proceedings of the
  AAAI Conference on Artificial Intelligence. vol.~35, pp. 5887--5895 (2021)

\bibitem{chen2023mld}
Chen, X., Jiang, B., Liu, W., Huang, Z., Fu, B., Chen, T., Yu, G.: Executing
  your commands via motion diffusion in latent space. In: Proceedings of the
  IEEE/CVF Conference on Computer Vision and Pattern Recognition. pp.
  18000--18010 (2023)

\bibitem{Dabral_2023mofusion}
Dabral, R., Mughal, M.H., Golyanik, V., Theobalt, C.: Mofusion: A framework for
  denoising-diffusion-based motion synthesis. In: Proceedings of the IEEE/CVF
  Conference on Computer Vision and Pattern Recognition (CVPR). pp. 9760--9770
  (June 2023)

\bibitem{diller2023cg}
Diller, C., Dai, A.: Cg-hoi: Contact-guided 3d human-object interaction
  generation. arXiv preprint arXiv:2311.16097  (2023)

\bibitem{ghosh2022imos}
Ghosh, A., Dabral, R., Golyanik, V., Theobalt, C., Slusallek, P.: {IMoS}:
  Intent-driven full-body motion synthesis for human-object interactions. arXiv
  preprint arXiv:2212.07555  (2022)

\bibitem{gong2023tm2d}
Gong, K., Lian, D., Chang, H., Guo, C., Jiang, Z., Zuo, X., Mi, M.B., Wang, X.:
  Tm2d: Bimodality driven 3{D} dance generation via music-text integration. In:
  Proceedings of the IEEE/CVF International Conference on Computer Vision. pp.
  9942--9952 (2023)

\bibitem{guo2022T2M}
Guo, C., Zou, S., Zuo, X., Wang, S., Ji, W., Li, X., Cheng, L.: Generating
  diverse and natural 3{D} human motions from text. In: Proceedings of the
  IEEE/CVF Conference on Computer Vision and Pattern Recognition. pp.
  5152--5161 (2022)

\bibitem{guo2022tm2t}
Guo, C., Zuo, X., Wang, S., Cheng, L.: Tm2t: Stochastic and tokenized modeling
  for the reciprocal generation of 3{D} human motions and texts. In: European
  Conference on Computer Vision. pp. 580--597. Springer (2022)

\bibitem{han2023chorus}
Han, S., Joo, H.: {CHORUS}: Learning canonicalized 3{D} human-object spatial
  relations from unbounded synthesized images. In: Proceedings of the IEEE/CVF
  International Conference on Computer Vision. pp. 15835--15846 (2023)

\bibitem{hassan2021posa}
Hassan, M., Ghosh, P., Tesch, J., Tzionas, D., Black, M.J.: Populating 3{D}
  scenes by learning human-scene interaction. In: Proceedings of the IEEE/CVF
  Conference on Computer Vision and Pattern Recognition. pp. 14708--14718
  (2021)

\bibitem{hassan2023physinter}
Hassan, M., Guo, Y., Wang, T., Black, M., Fidler, S., Peng, X.B.: Synthesizing
  physical character-scene interactions. arXiv preprint arXiv:2302.00883
  (2023)

\bibitem{ho2020ddpm}
Ho, J., Jain, A., Abbeel, P.: Denoising diffusion probabilistic models.
  Advances in neural information processing systems  \textbf{33},  6840--6851
  (2020)

\bibitem{huang2023scenediffuser}
Huang, S., Wang, Z., Li, P., Jia, B., Liu, T., Zhu, Y., Liang, W., Zhu, S.C.:
  Diffusion-based generation, optimization, and planning in 3{D} scenes. In:
  Proceedings of the IEEE/CVF Conference on Computer Vision and Pattern
  Recognition. pp. 16750--16761 (2023)

\bibitem{huang2022intercap}
Huang, Y., Taheri, O., Black, M.J., Tzionas, D.: Inter{C}ap: Joint markerless
  3{D} tracking of humans and objects in interaction. In: DAGM German
  Conference on Pattern Recognition. pp. 281--299. Springer (2022)

\bibitem{ijcai2023stackflow}
Huo, C., Shi, Y., Ma, Y., Xu, L., Yu, J., Wang, J.: Stackflow: Monocular
  human-object reconstruction by stacked normalizing flow with offset. In:
  Elkind, E. (ed.) Proceedings of the Thirty-Second International Joint
  Conference on Artificial Intelligence, {IJCAI-23}. pp. 902--910.
  International Joint Conferences on Artificial Intelligence Organization (8
  2023). \doi{10.24963/ijcai.2023/100},
  \url{https://doi.org/10.24963/ijcai.2023/100}, main Track

\bibitem{jiang2023chairs}
Jiang, N., Liu, T., Cao, Z., Cui, J., Zhang, Z., Chen, Y., Wang, H., Zhu, Y.,
  Huang, S.: Full-body articulated human-object interaction. In: Proceedings of
  the IEEE/CVF International Conference on Computer Vision. pp. 9365--9376
  (2023)

\bibitem{kim2023flame}
Kim, J., Kim, J., Choi, S.: Flame: Free-form language-based motion synthesis \&
  editing. In: Proceedings of the AAAI Conference on Artificial Intelligence.
  vol.~37, pp. 8255--8263 (2023)

\bibitem{kulkarni2023nifty}
Kulkarni, N., Rempe, D., Genova, K., Kundu, A., Johnson, J., Fouhey, D.,
  Guibas, L.: Nifty: Neural object interaction fields for guided human motion
  synthesis. arXiv preprint arXiv:2307.07511  (2023)

\bibitem{li2022danceformer}
Li, B., Zhao, Y., Zhelun, S., Sheng, L.: Danceformer: Music conditioned 3{D}
  dance generation with parametric motion transformer. In: Proceedings of the
  AAAI Conference on Artificial Intelligence. vol.~36, pp. 1272--1279 (2022)

\bibitem{li2023controllable}
Li, J., Clegg, A., Mottaghi, R., Wu, J., Puig, X., Liu, C.K.: Controllable
  human-object interaction synthesis. arXiv preprint arXiv:2312.03913  (2023)

\bibitem{li2023egoego}
Li, J., Liu, K., Wu, J.: Ego-body pose estimation via ego-head pose estimation.
  In: Proceedings of the IEEE/CVF Conference on Computer Vision and Pattern
  Recognition. pp. 17142--17151 (2023)

\bibitem{li2023OMOMO}
Li, J., Wu, J., Liu, C.K.: Object motion guided human motion synthesis. arXiv
  preprint arXiv:2309.16237  (2023)

\bibitem{li2021choreographer}
Li, R., Yang, S., Ross, D.A., Kanazawa, A.: Ai choreographer: Music conditioned
  3{D} dance generation with aist++. In: Proceedings of the IEEE/CVF
  International Conference on Computer Vision. pp. 13401--13412 (2021)

\bibitem{liang2023intergen}
Liang, H., Zhang, W., Li, W., Yu, J., Xu, L.: Intergen: Diffusion-based
  multi-human motion generation under complex interactions. arXiv preprint
  arXiv:2304.05684  (2023)

\bibitem{lim2023mammos}
Lim, D., Jeong, C., Kim, Y.M.: {MAMMOS}: Mapping multiple human motion with
  scene understanding and natural interactions. In: Proceedings of the IEEE/CVF
  International Conference on Computer Vision. pp. 4278--4287 (2023)

\bibitem{liu2018basketball}
Liu, L., Hodgins, J.: Learning basketball dribbling skills using trajectory
  optimization and deep reinforcement learning. ACM Transactions on Graphics
  (TOG)  \textbf{37}(4),  1--14 (2018)

\bibitem{loper2015smpl}
Loper, M., Mahmood, N., Romero, J., Pons-Moll, G., Black, M.J.: Smpl: a skinned
  multi-person linear model. ACM Transactions on Graphics (TOG)
  \textbf{34}(6),  1--16 (2015)

\bibitem{lucas2022posegpt}
Lucas, T., Baradel, F., Weinzaepfel, P., Rogez, G.: Posegpt: Quantization-based
  3{D} human motion generation and forecasting. In: European Conference on
  Computer Vision. pp. 417--435. Springer (2022)

\bibitem{merel2020catchcarry}
Merel, J., Tunyasuvunakool, S., Ahuja, A., Tassa, Y., Hasenclever, L., Pham,
  V., Erez, T., Wayne, G., Heess, N.: Catch \& carry: reusable neural
  controllers for vision-guided whole-body tasks. ACM Transactions on Graphics
  (TOG)  \textbf{39}(4),  39--1 (2020)

\bibitem{pavlakos2019smplx}
Pavlakos, G., Choutas, V., Ghorbani, N., Bolkart, T., Osman, A.A., Tzionas, D.,
  Black, M.J.: Expressive body capture: 3d hands, face, and body from a single
  image. In: Proceedings of the IEEE/CVF conference on computer vision and
  pattern recognition. pp. 10975--10985 (2019)

\bibitem{peng2023hoi}
Peng, X., Xie, Y., Wu, Z., Jampani, V., Sun, D., Jiang, H.: Hoi-diff:
  Text-driven synthesis of 3d human-object interactions using diffusion models.
  arXiv preprint arXiv:2312.06553  (2023)

\bibitem{petrov2023objectpopup}
Petrov, I.A., Marin, R., Chibane, J., Pons-Moll, G.: Object pop-up: Can we
  infer 3{D} objects and their poses from human interactions alone? In:
  Proceedings of the IEEE/CVF Conference on Computer Vision and Pattern
  Recognition. pp. 4726--4736 (2023)

\bibitem{petrovich2021action}
Petrovich, M., Black, M.J., Varol, G.: Action-conditioned 3{D} human motion
  synthesis with transformer {VAE}. In: Proceedings of the IEEE/CVF
  International Conference on Computer Vision. pp. 10985--10995 (2021)

\bibitem{Petrovich_2021_Actor}
Petrovich, M., Black, M.J., Varol, G.: Action-conditioned 3{D} human motion
  synthesis with transformer vae. In: Proceedings of the IEEE/CVF International
  Conference on Computer Vision (ICCV). pp. 10985--10995 (October 2021)

\bibitem{petrovich2022temos}
Petrovich, M., Black, M.J., Varol, G.: Temos: Generating diverse human motions
  from textual descriptions. In: European Conference on Computer Vision. pp.
  480--497. Springer (2022)

\bibitem{pi2023hierarchical}
Pi, H., Peng, S., Yang, M., Zhou, X., Bao, H.: Hierarchical generation of
  human-object interactions with diffusion probabilistic models. In:
  Proceedings of the IEEE/CVF International Conference on Computer Vision. pp.
  15061--15073 (2023)

\bibitem{qi2017pointnet}
Qi, C.R., Su, H., Mo, K., Guibas, L.J.: Pointnet: Deep learning on point sets
  for 3{D} classification and segmentation. In: Proceedings of the IEEE
  conference on computer vision and pattern recognition. pp. 652--660 (2017)

\bibitem{radford2021CLIP}
Radford, A., Kim, J.W., Hallacy, C., Ramesh, A., Goh, G., Agarwal, S., Sastry,
  G., Askell, A., Mishkin, P., Clark, J., et~al.: Learning transferable visual
  models from natural language supervision. In: International conference on
  machine learning. pp. 8748--8763. PMLR (2021)

\bibitem{ramesh2022hierarchicaltext}
Ramesh, A., Dhariwal, P., Nichol, A., Chu, C., Chen, M.: Hierarchical
  text-conditional image generation with clip latents. arXiv preprint
  arXiv:2204.06125  \textbf{1}(2), ~3 (2022)

\bibitem{razali2023acgmanipulation}
Razali, H., Demiris, Y.: Action-conditioned generation of bimanual object
  manipulation sequences. In: Proceedings of the AAAI Conference on Artificial
  Intelligence. vol.~37, pp. 2146--2154 (2023)

\bibitem{ren2023diffmotion}
Ren, Z., Pan, Z., Zhou, X., Kang, L.: Diffusion motion: Generate text-guided
  3{D} human motion by diffusion model. In: ICASSP 2023-2023 IEEE International
  Conference on Acoustics, Speech and Signal Processing (ICASSP). pp.~1--5.
  IEEE (2023)

\bibitem{MANO:SIGGRAPHASIA:2017}
Romero, J., Tzionas, D., Black, M.J.: Embodied hands: Modeling and capturing
  hands and bodies together. ACM Transactions on Graphics, (Proc. SIGGRAPH
  Asia)  \textbf{36}(6) (Nov 2017)

\bibitem{shafir2023priorMDM}
Shafir, Y., Tevet, G., Kapon, R., Bermano, A.H.: Human motion diffusion as a
  generative prior. arXiv preprint arXiv:2303.01418  (2023)

\bibitem{starke2019nsm}
Starke, S., Zhang, H., Komura, T., Saito, J.: Neural state machine for
  character-scene interactions. ACM Trans. Graph.  \textbf{38}(6),  209--1
  (2019)

\bibitem{taheri2021goal}
Taheri, O., Choutas, V., Black, M.J., Tzionas, D.: {GOAL}: {G}enerating {4D}
  whole-body motion for hand-object grasping. In: Conference on Computer Vision
  and Pattern Recognition ({CVPR}) (2022), \url{https://goal.is.tue.mpg.de}

\bibitem{taheri2023grip}
Taheri, O., Zhou, Y., Tzionas, D., Zhou, Y., Ceylan, D., Pirk, S., Black, M.J.:
  {GRIP}: Generating interaction poses using latent consistency and spatial
  cues. arXiv preprint arXiv:2308.11617  (2023)

\bibitem{Tanaka_2023_RAIG}
Tanaka, M., Fujiwara, K.: Role-aware interaction generation from textual
  description. In: Proceedings of the IEEE/CVF International Conference on
  Computer Vision (ICCV). pp. 15999--16009 (October 2023)

\bibitem{tevet2022motionclip}
Tevet, G., Gordon, B., Hertz, A., Bermano, A.H., Cohen-Or, D.: Motionclip:
  Exposing human motion generation to clip space. In: European Conference on
  Computer Vision. pp. 358--374. Springer (2022)

\bibitem{tevet2023MDM}
Tevet, G., Raab, S., Gordon, B., Shafir, Y., Cohen-or, D., Bermano, A.H.: Human
  motion diffusion model. In: The Eleventh International Conference on Learning
  Representations (2023), \url{https://openreview.net/forum?id=SJ1kSyO2jwu}

\bibitem{Wang_2023_fgt2m}
Wang, Y., Leng, Z., Li, F.W.B., Wu, S.C., Liang, X.: Fg-{T}2{M}: Fine-grained
  text-driven human motion generation via diffusion model. In: Proceedings of
  the IEEE/CVF International Conference on Computer Vision (ICCV). pp.
  22035--22044 (October 2023)

\bibitem{wang2022humanise}
Wang, Z., Chen, Y., Liu, T., Zhu, Y., Liang, W., Huang, S.: Humanise:
  Language-conditioned human motion generation in 3{D} scenes. Advances in
  Neural Information Processing Systems  \textbf{35},  14959--14971 (2022)

\bibitem{wu2022saga}
Wu, Y., Wang, J., Zhang, Y., Zhang, S., Hilliges, O., Yu, F., Tang, S.: Saga:
  Stochastic whole-body grasping with contact. In: European Conference on
  Computer Vision. pp. 257--274. Springer (2022)

\bibitem{xiao2023unihsi}
Xiao, Z., Wang, T., Wang, J., Cao, J., Zhang, W., Dai, B., Lin, D., Pang, J.:
  Unified human-scene interaction via prompted chain-of-contacts. arXiv
  preprint arXiv:2309.07918  (2023)

\bibitem{xie2022chore}
Xie, X., Bhatnagar, B.L., Pons-Moll, G.: Chore: Contact, human and object
  reconstruction from a single rgb image. In: European Conference on Computer
  Vision. pp. 125--145. Springer (2022)

\bibitem{xie2023vistracker}
Xie, X., Bhatnagar, B.L., Pons-Moll, G.: Visibility aware human-object
  interaction tracking from single rgb camera. In: Proceedings of the IEEE/CVF
  Conference on Computer Vision and Pattern Recognition. pp. 4757--4768 (2023)

\bibitem{xie2023omnicontrol}
Xie, Y., Jampani, V., Zhong, L., Sun, D., Jiang, H.: Omnicontrol: Control any
  joint at any time for human motion generation. arXiv preprint
  arXiv:2310.08580  (2023)

\bibitem{xie2023boxmani}
Xie, Z., Tseng, J., Starke, S., van~de Panne, M., Liu, C.K.: Hierarchical
  planning and control for box loco-manipulation. arXiv preprint
  arXiv:2306.09532  (2023)

\bibitem{Xu_2023_actformer}
Xu, L., Song, Z., Wang, D., Su, J., Fang, Z., Ding, C., Gan, W., Yan, Y., Jin,
  X., Yang, X., Zeng, W., Wu, W.: Act{F}ormer: A gan-based transformer towards
  general action-conditioned 3{D} human motion generation. In: Proceedings of
  the IEEE/CVF International Conference on Computer Vision (ICCV). pp.
  2228--2238 (October 2023)

\bibitem{xu2023interdiff}
Xu, S., Li, Z., Wang, Y.X., Gui, L.Y.: Interdiff: Generating 3{D} human-object
  interactions with physics-informed diffusion. In: Proceedings of the IEEE/CVF
  International Conference on Computer Vision. pp. 14928--14940 (2023)

\bibitem{yang2023coherentsamping}
Yang, Z., Su, B., Wen, J.R.: Synthesizing long-term human motions with
  diffusion models via coherent sampling. arXiv preprint arXiv:2308.01850
  (2023)

\bibitem{Ye2022summon}
Ye, S., Wang, Y., Li, J., Park, D., Liu, C.K., Xu, H., Wu, J.: Scene synthesis
  from human motion. In: SIGGRAPH Asia 2022 Conference Papers. SA '22,
  Association for Computing Machinery, New York, NY, USA (2022).
  \doi{10.1145/3550469.3555426}, \url{https://doi.org/10.1145/3550469.3555426}

\bibitem{yi2023mime}
Yi, H., Huang, C.H.P., Tripathi, S., Hering, L., Thies, J., Black, M.J.: Mime:
  Human-aware 3{D} scene generation. In: Proceedings of the IEEE/CVF Conference
  on Computer Vision and Pattern Recognition. pp. 12965--12976 (2023)

\bibitem{Yi_2022_mover}
Yi, H., Huang, C.H.P., Tzionas, D., Kocabas, M., Hassan, M., Tang, S., Thies,
  J., Black, M.J.: Human-aware object placement for visual environment
  reconstruction. In: Proceedings of the IEEE/CVF Conference on Computer Vision
  and Pattern Recognition (CVPR). pp. 3959--3970 (June 2022)

\bibitem{yuan2023physdiff}
Yuan, Y., Song, J., Iqbal, U., Vahdat, A., Kautz, J.: Physdiff: Physics-guided
  human motion diffusion model. In: Proceedings of the IEEE/CVF International
  Conference on Computer Vision. pp. 16010--16021 (2023)

\bibitem{zhang2020phosa}
Zhang, J.Y., Pepose, S., Joo, H., Ramanan, D., Malik, J., Kanazawa, A.:
  Perceiving 3{D} human-object spatial arrangements from a single image in the
  wild. In: European Conference on Computer Vision (ECCV) (2020)

\bibitem{zhang2023t2mgpt}
Zhang, J., Zhang, Y., Cun, X., Huang, S., Zhang, Y., Zhao, H., Lu, H., Shen,
  X.: T2m-gpt: Generating human motion from textual descriptions with discrete
  representations. arXiv preprint arXiv:2301.06052  (2023)

\bibitem{zhang2023neuraldome}
Zhang, J., Luo, H., Yang, H., Xu, X., Wu, Q., Shi, Y., Yu, J., Xu, L., Wang,
  J.: Neural{D}ome: A neural modeling pipeline on multi-view human-object
  interactions. In: Proceedings of the IEEE/CVF Conference on Computer Vision
  and Pattern Recognition. pp. 8834--8845 (2023)

\bibitem{Zhang_2023_remodiffuse}
Zhang, M., Guo, X., Pan, L., Cai, Z., Hong, F., Li, H., Yang, L., Liu, Z.:
  Re{M}o{D}iffuse: Retrieval-augmented motion diffusion model. In: Proceedings
  of the IEEE/CVF International Conference on Computer Vision (ICCV). pp.
  364--373 (October 2023)

\bibitem{zhang2022couch}
Zhang, X., Bhatnagar, B.L., Starke, S., Guzov, V., Pons-Moll, G.: Couch:
  Towards controllable human-chair interactions. In: European Conference on
  Computer Vision. pp. 518--535. Springer (2022)

\bibitem{zhao2022compositional}
Zhao, K., Wang, S., Zhang, Y., Beeler, T., Tang, S.: Compositional human-scene
  interaction synthesis with semantic control. In: European Conference on
  Computer Vision. pp. 311--327. Springer (2022)

\bibitem{Zhao2023DIMOS}
Zhao, K., Zhang, Y., Wang, S., Beeler, T., , Tang, S.: Synthesizing diverse
  human motions in 3{D} indoor scenes. In: International conference on computer
  vision (ICCV) (2023)

\bibitem{zhong2023attt2m}
Zhong, C., Hu, L., Zhang, Z., Xia, S.: Attt2m: Text-driven human motion
  generation with multi-perspective attention mechanism. In: Proceedings of the
  IEEE/CVF International Conference on Computer Vision. pp. 509--519 (2023)

\bibitem{zhou2022toch}
Zhou, K., Bhatnagar, B.L., Lenssen, J.E., Pons-Moll, G.: Toch: Spatio-temporal
  object-to-hand correspondence for motion refinement. In: European Conference
  on Computer Vision. pp. 1--19. Springer (2022)

\bibitem{zhuang2022music2dance}
Zhuang, W., Wang, C., Chai, J., Wang, Y., Shao, M., Xia, S.: Music2dance:
  Dancenet for music-driven dance generation. ACM Transactions on Multimedia
  Computing, Communications, and Applications (TOMM)  \textbf{18}(2),  1--21
  (2022)

\end{thebibliography}

\clearpage

\appendix


\setcounter{page}{1}
\setcounter{table}{0}   
\setcounter{figure}{0}

\renewcommand{\thetable}{A\arabic{table}}
\renewcommand{\thefigure}{A\arabic{figure}}

Supplementary materials includes additional details of our Text-BEHAVE dataset, experiment results and evaluation model. Sec.~\ref{appdix:dataset} showcases the object models in our dataset and describes the details of annotation. Sec.~\ref{appdix:results} provides more experiment results on out-of-the-dataset generation, guidance scale analysis and object conditioned generation. Sec.~\ref{appdix:eval} gives training details of the evaluation model. Sec.~\ref{appdix:implementation} adds other implementation details and Sec.~\ref{appdix:failure} shows some failure cases.
\section{Data analysis}
\label{appdix:dataset}
\subsection{Object Categories}
\begin{figure}[htp]
    \centering
    \includegraphics[width=0.75\textwidth]{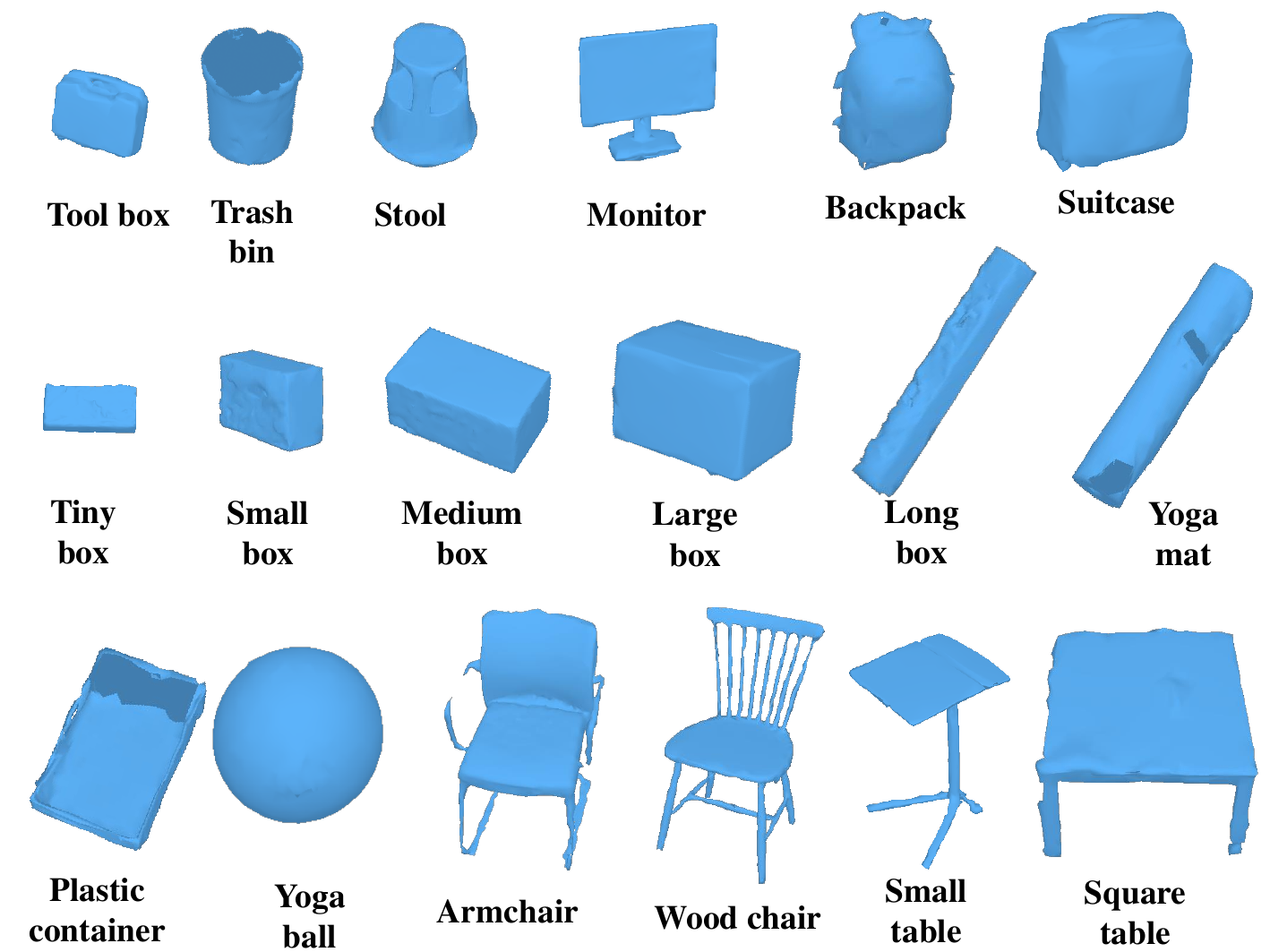}
    \caption{Object models in Text-BEHAVE.}
    \label{fig:object_models}
\end{figure}

 Text-BEHAVE comprises 18 object models, for which the original dataset furnishes comprehensive motion data captured at 30 fps. As depicted in Figure~\ref{fig:object_models}, these models span diverse categories, sizes, and geometries.
\begin{figure}
    \centering
    \includegraphics[width=0.8\textwidth]{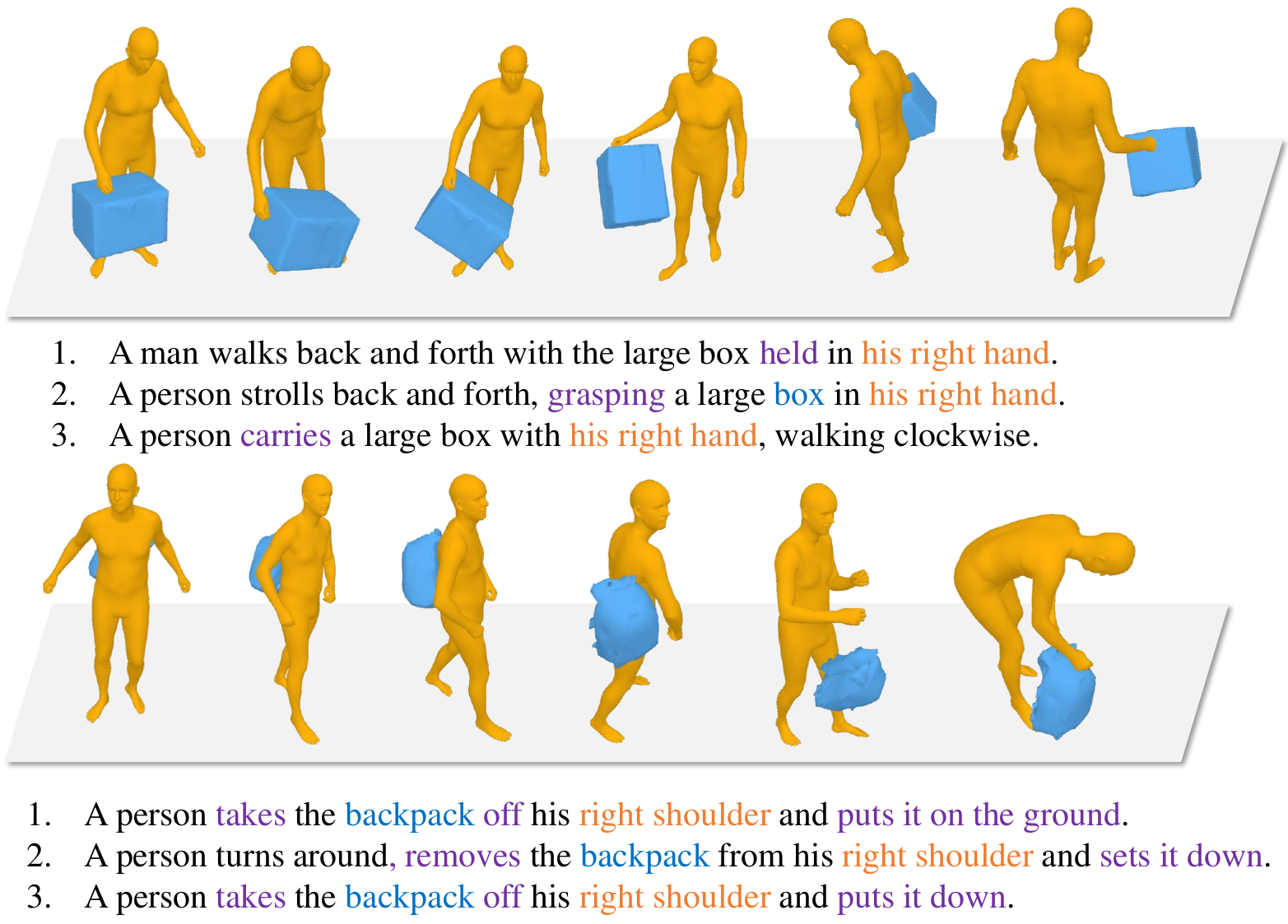}
    \caption{Examples in Text-BEHAVE dataset. Each interactions segment is annoted with 3 textual descriptions.}
    \label{fig:text_annotation}
\end{figure}
\subsection{Details of Text Annotation}
We provide a instruction for the Crowd-sourcing Platform to desribe the interactions in the videos of BEHAVE \cite{bhatnagar2022behave}, and then manually replace the object name. Since origin BEHAVE dataset includes amounts of non-interactive and redundant actions in a long motion sequence, we ask the annotators to divide the long sequence and mark out the meaningless segments. A detailed instruction and textual example are as the follows,

\noindent\textbf{Annotation Requirement}:
\begin{itemize}
    \item Divide the entire sequence into multiple interaction segments, ensuring each segment has complete semantics, and keep the duration within 3 to 10 seconds.
    \item Each interaction description should include \textbf{character's action}, \textbf{object name}, and try to include \textbf{body part} in contact with the object \textbf{changes in the object's position} and \textbf{changes in the person's position}.
    \item For long videos, provide both a summary description of the entire video and segmented descriptions.
\end{itemize}

\noindent\textbf{Example Segments}:
\begin{enumerate}
    \item `A person bends down to pick up a backpack from the ground in front'.
    \item `Holding the bottom of the backpack with both hands, he walks counterclockwise'.
    \item `He lifts the backpack onto his left shoulder, with his left hand supporting the bottom, walking clockwise'.\
    \item `He shifts the backpack from his left shoulder to the right shoulder, starting to walk counterclockwise'.
    \item `Using both hands, he raises the backpack above his head'.
    \item `With both hands, he moves the backpack from above his head to the front of his chest'.
    \item `He walks, holding the backpack with both hands, pressed against his abdomen'.
\end{enumerate}
Each interaction segment is annotated with 3 textual descriptions, as shown in Figure~\ref{fig:text_annotation}.

\section{Additional Implementation Details}
\label{appdix:implementation}
Transformer layers in early generation are with hidden size 512, feed forward size 1024, dropout 0.1 and `gelu' activation. Transformer layers in Position encoder and Rotation encoder are with hidden size 156, feed forward size 128, dropout 0.1 and `gelu' activation. Loss weights are $\lambda_{rel}=0.01$, $\lambda_{dist}=0.01$ and $\lambda_{vel}=0.0003$ respectively. Guidance scale $s$ is set to 2.5 when sampling the interactions. The total training epochs are 200.

\section{More Results}
\label{appdix:results}

\subsection{Effectiveness of Guidance Scale}

In Table~\ref{tab:gs}, we test different guidance scales and show the effectiveness of large guidance scale. 

\begin{table}[h]
	\begin{center}
		\centering
		\caption{\textbf{Quantitative comparisons of different guidance scales on the Text-BEHAVE test set for Text2HOI generation.} All the evaluations run 20 times. $\pm$ indicates the 95\% confidence interval. \textbf{Bold} indicates best result.}  
  \label{tab:gs}
\resizebox{\textwidth}{!}{		
		\begin{tabular}{lccccccc}
			\toprule
			\multirow{2}{*}{Guidance scale}  & \multicolumn{3}{c}{R Precision$\uparrow$} & \multirow{2}{*}{FID $\downarrow$} & \multirow{2}{*}{MM Dist$\downarrow$}  & \multirow{2}{*}{Diversity$\rightarrow $} & \multirow{2}{*}{MModality $\uparrow$} \\
			\cmidrule(lr){2-4}
			& Top 1 & Top 2  & Top 3 \\
			\midrule
			\textbf{Real motions}    &  $0.994^{\pm .003}$ &  $0.998^{\pm .001}$  & $1.000^{\pm .000}$   &  $0.031^{\pm .002}$ &  $3.377^{\pm .004}$  & $22.832^{\pm .074}$   & $-$  \\
			\midrule
			0.0  &  $0.094^{\pm .007}$ &  $0.183^{\pm .014}$  & $0.232^{\pm .011}$   &  $2.742^{\pm .033}$ &  $7.428^{\pm .018}$  & $22.271^{\pm .114}$   & $1.612^{\pm .123}$  \\
			0.5   &  $0.181^{\pm .010}$ &  $0.277^{\pm .014}$  & $0.353^{\pm .015}$   &   $2.493^{\pm .060}$&  $7.057^{\pm .030}$  & ${22.533}^{\pm .110}$   & $\textbf{3.781}^{\pm .205}$ \\          
			1.0   &  $0.225^{\pm .008}$ &  $0.328^{\pm .009}$  & $0.404^{\pm .009}$   &   $2.133^{\pm .021}$&  $6.907^{\pm .012}$  & ${22.676}^{\pm .088}$   & $3.375^{\pm .093}$ \\

   			1.5   &  $0.224^{\pm .009}$ &  $0.328^{\pm .009}$  & $0.399^{\pm .010}$   &  $2.033^{\pm .023}$ &  $7.008^{\pm .018}$  & $22.705^{\pm .141}$ & $2.717^{\pm .109}$ \\
            2.0    &  $0.249^{\pm .009}$ &  $0.358^{\pm .010}$  & $0.409^{\pm .003}$   &  $2.048^{\pm .021}$ &  $6.907^{\pm .014}$  & $22.687^{\pm .078}$ & $2.535^{\pm .109}$ \\
            2.5    &  $\textbf{0.250}^{\pm .009}$ &  $\textbf{0.362}^{\pm .008}$  & $0.411^{\pm .001}$   &  $\textbf{1.989}^{\pm .020}$ &  $\textbf{6.874}^{\pm .014}$  & $22.701^{\pm .081}$ & $2.575^{\pm .071}$ \\
         	3.0   &  $0.224^{\pm .009}$ &  $0.334^{\pm .011}$  & $0.408^{\pm .010   }$   &  $2.066^{\pm .016}$ &  $6.905^{\pm .010}$  & $\textbf{22.764}^{\pm .058}$ & $2.492^{\pm .010}$ \\
         	4.0   &  $0.231^{\pm .008}$ &  $0.348^{\pm .008}$  & $\textbf{0.426}^{\pm .007}$  &  $2.081^{\pm .018}$ &  $6.883^{\pm .011}$  & $22.655^{\pm .080}$ & $2.513^{\pm .063}$ \\

			\bottomrule
		\end{tabular}
  }
	\end{center}

\end{table}

\subsection{Out-of-the-dataset Generation}
\begin{figure}[htp]
    \centering
    \includegraphics[width=0.65\textwidth]{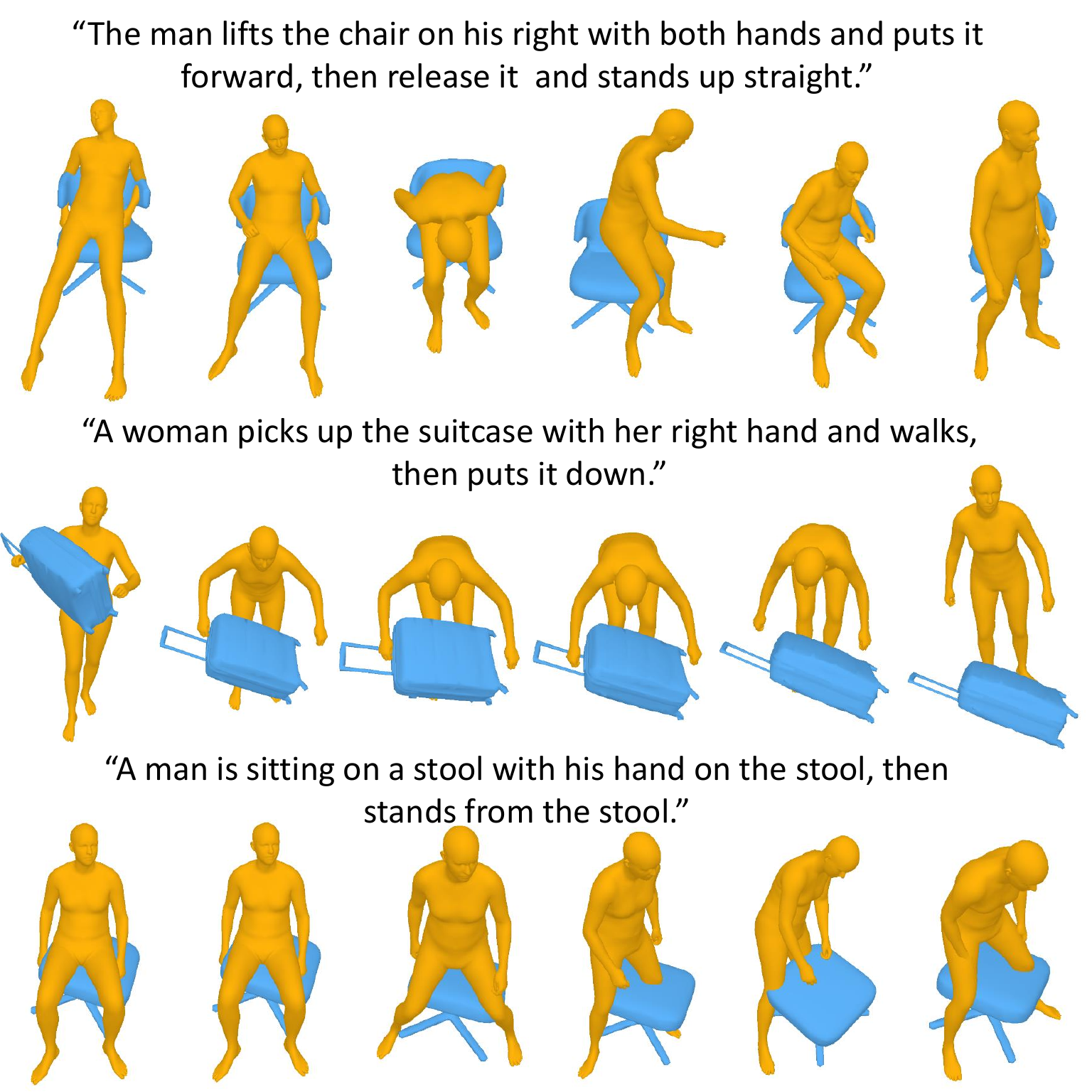}
    \caption{We also test our model on the InterCap \cite{huang2022intercap}. Even if the object shape is not in Text-BEHAVE, our model demonstrates the ability of generalization with respect to the object shape.}
    \label{fig:intercap}
\end{figure}

We additionally annotate the InterCap dataset \cite{huang2022intercap}. However, due to its scale and the transition cost from SMPL-X \cite{pavlakos2019smplx} to SMPL-H \cite{MANO:SIGGRAPHASIA:2017}, we opt to selectively include its object models for supplementary testing, particularly in scenarios involving out-of-the-dataset generation. Notably, our model demonstrates proficiency in generating interactions with objects that are beyond the scope of the training data, as in Figure~\ref{fig:intercap}.

\subsection{Motion Reconstruction}
Motion reconstruction metrics using the Best-of-many \cite{bhattacharyya2018bestofmany} method only provides a marginal reference for possible best result over the test set \cite{xu2023interdiff, li2023OMOMO}, as in Tab.~\ref{tab:recon}. Through 20 times repeated generation under the same text, THOR can generate interactions mostly close to the ground truth, especially for the object motion. It also turns out the human motion can also be refined during the diffusion process with intervention and intervention losses. \textit{MPJPE} refers to mean per-joint position error, \emph{$\text{v2v}_h$} and \emph{$\text{v2v}_o$} refers to vextex position error of human and object \cite{zhang2023neuraldome}, \emph{Rot. Err.} refers to the Frobenius norm of rotation matrix difference and \emph{Tr. Err.} is the translation error of object \cite{li2023OMOMO}.
\begin{table}
  \centering
  \caption{\textbf{Best-of-many} Motion reconstruction metrics. All the evaluations run 20 times and the best results are selected.}
  \resizebox{0.7\textwidth}{!}{	
  \begin{tabular}{@{}l@{}ccccc@{}}
    \toprule
    \multirow{2}{*}{Methods} & \multicolumn{2}{c}{Human} & \multicolumn{3}{c}{Object} \\
    \cmidrule(lr){2-3}
    \cmidrule(lr){4-6}
        & MPJPE$\downarrow$  &  $\text{v2v}_h$ $\downarrow$   & Rot. Err. $\downarrow$  & Tr. Err.$\downarrow$  &  $\text{v2v}_o$ $\downarrow$  \\
    \midrule
    Text-MDM &60.60 & 56.30 &1.80 & 53.82 & 60.03\\
    \midrule
    THOR w/o intervention &59.71 & 55.61 &1.76 &51.80 & 58.40\\
    THOR w/o dist. loss & \textbf{59.50} & \textbf{55.18} & 1.77 & 51.74 & 57.90 \\
    THOR w/o rel. loss & 59.85 & 55.56 & 1.80 & 51.79 & 58.09\\
    THOR (Full) & 59.69 & 55.44 & \textbf{1.73} & \textbf{51.25} & \textbf{57.71} \\
    \bottomrule
  \end{tabular}
  }
  \label{tab:recon}
\end{table}

\section{Evaluation Details}
\label{appdix:eval}
Our evaluation model follows \cite{tevet2022motionclip, liang2023intergen}, which is composed of a text feature extractor and human-object motion feature extractor. 
The text feature extractor comprises token embedding from CLIP \cite{radford2021CLIP} and a transformer encoder of 8 layers with 4 heads. The motion feature extractor is the same as in our model THOR, including a transformer encoder of 8 layers with 4 heads. Their hidden size is 512 and feed forward size is 1024. The latent dimension for text and motion are both $512$. It is trained on the entire Text-BEHAVE dataset with contrastive loss \cite{radford2021CLIP, tevet2022motionclip, liang2023intergen}:
\begin{equation}
    \begin{aligned}
        \mathcal{L}_{text} &= CrossEntropy(\boldsymbol z_{text}, \boldsymbol z_{cls}) \\
        \mathcal{L}_{motion} &= CrossEntropy(\boldsymbol z_{motion}, \boldsymbol z_{cls}) \\
        \mathcal{L} &= \frac{1}{2}(\mathcal{L}_{text} + \mathcal{L}_{motion}) 
    \end{aligned}
\end{equation}
where $\boldsymbol z_{text}$ and $\boldsymbol z_{motion}$ are extracted text and motion features, and $\boldsymbol z_{cls}$ are labels following \cite{radford2021CLIP}.
The evaluation model is trained with batch size of $64$ and learning rate $1e^{-4}$ with weight decay $1e^{-4}$. 
\begin{figure}[ht]
    \centering
    \includegraphics[width=0.65\textwidth]{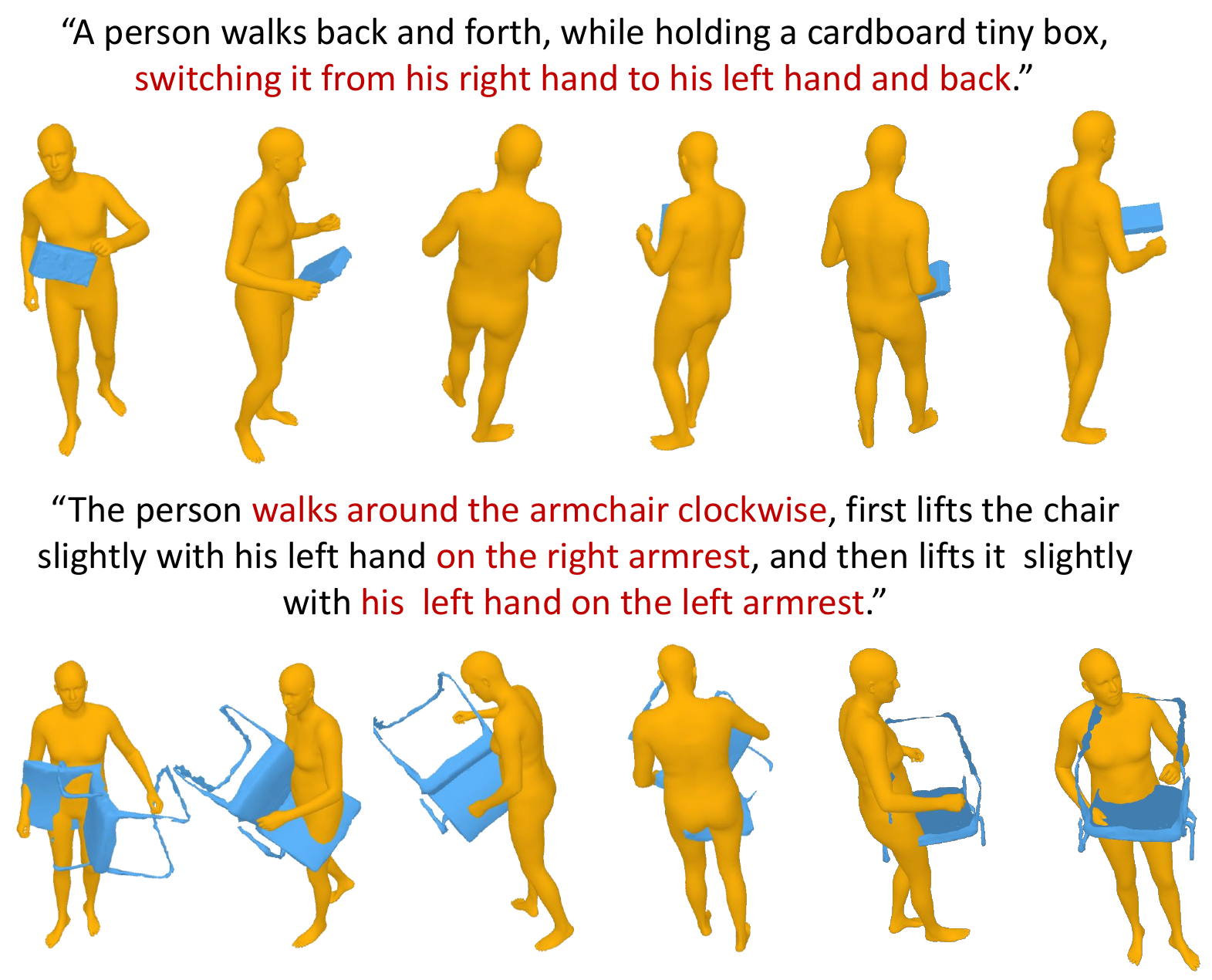}
    \caption{Examples of failure cases. It mainly arises from discrepancies between complex text prompts and penetrations posed by objects with intricate geometries.}
    \label{fig:failure_case}
\end{figure}
\section{Failure Cases}
\label{appdix:failure}

Text2HOI is a challenging task, and there are failed generation results as well. Some failure examples are shown in Figuire~\ref{fig:failure_case}. Floating movements of objects and penetrations between humans and objects may occur, particularly when dealing with objects that are excessively small or possess an exceedingly complex structure. Besides, there are failures dealing with the long-term textual descriptions, leading to missing interactions or transitions.

\end{document}